\documentclass{article}
\usepackage{hyphenat}
\PassOptionsToPackage{numbers, compress}{natbib}
\usepackage[preprint]{neurips_2023}

\usepackage[utf8]{inputenc} 
\usepackage[T1]{fontenc}    
\usepackage{hyperref}       
\usepackage{url}            
\usepackage{booktabs}       
\usepackage{tabularx}
\usepackage{amsfonts}       
\usepackage{nicefrac}       
\usepackage{microtype}      
\usepackage{xcolor}         
\usepackage{amsmath}
\usepackage{array}
\usepackage{amssymb}
\usepackage{mathtools}
\usepackage{amsthm}
\usepackage{multirow}
\usepackage{wrapfig}
\usepackage[capitalize,noabbrev]{cleveref}
\usepackage{graphicx}
\usepackage{subfigure} 
\usepackage{float}
\usepackage{algorithm} 
\usepackage{algorithmic}
\usepackage{textcomp} 
\usepackage{listings} 
\usepackage{enumitem}
\usepackage{utfsym}
\usepackage{times}
\usepackage{latexsym}
\usepackage{fancyvrb}
\usepackage{tablefootnote}
\usepackage{CJKutf8}

\newcommand{\commentout}[1]{}

\title{FireRedAudio: A General-Purpose Audio Language Model with Decoupled Continuous Representations for Understanding and Generation}

\author{
FireRed Team\\
Xiaohongshu
}

\begin{document}
\maketitle

\begin{abstract}
A unified audio model must recognize and understand linguistic, paralinguistic, and environmental information while supporting speech synthesis and editing. A key challenge is representation: understanding favors compact features suited to long-context modeling, whereas speech generation requires reconstructible features that preserve fine-grained acoustic detail. We introduce FireRedAudio, a general-purpose audio language model with a shared 9B-parameter LLM. To the best of our knowledge, it is the first publicly disclosed unified audio-language model to provide separate continuous input representations for understanding and generation within a single trainable autoregressive LLM. Audio to be recognized or analyzed is processed by a dedicated Audio Encoder, while speech inputs for generation use a RedAE-based pathway. The LLM directly generates text or conditions a flow-matching DiT to produce continuous acoustic latents. Through progressive multitask training, FireRedAudio supports ASR and audio understanding, with the latter extending to recordings of up to one hour, as well as zero-shot TTS, Instruct TTS, and semantic and acoustic speech editing. Its structured organization of long-form audio achieves second-level timestamp accuracy. Across comprehensive evaluations, FireRedAudio achieves competitive or leading performance in audio understanding and multilingual ASR, strong content accuracy and speaker preservation in zero-shot TTS, leading instruction following in Instruct TTS, and substantial improvements over Ming-UniAudio-Edit in both semantic and acoustic speech editing. These results demonstrate the viability of decoupled continuous input representations for unifying audio understanding and continuous-latent speech generation in a model of moderate scale. Our code is available at \url{https://github.com/FireRedTeam/FireRedAudio}.
\end{abstract}

\section{Introduction}

General-purpose audio modeling spans automatic speech recognition (ASR), audio understanding, zero-shot text-to-speech synthesis (TTS), natural-language-controlled TTS (Instruct TTS), speech editing, and long-form audio analysis through a unified interface. A common language backbone allows these tasks to reuse linguistic knowledge, instruction-following ability, and reasoning. However, understanding and generation place different demands on audio representations. Understanding benefits from compact representations that organize task-relevant acoustic cues for long-context modeling, whereas speech generation requires waveform-reconstructible representations that preserve fine-grained details such as timbre, prosody, pitch, and speaking style. Requiring one representation to provide both task-oriented abstraction and faithful waveform reconstruction can therefore create a performance trade-off. Prior work~\cite{zhou2026f3} has also observed that continuous autoencoder latents optimized for waveform reconstruction tend to emphasize acoustic detail and may not exhibit the structure required for high-level understanding, whereas self-supervised audio representations designed for recognition and analysis generally lack a decoding interface for directly reconstructing high-fidelity waveforms.

A mainstream approach in existing unified audio-language models is to predict quantized audio tokens for generation while representing input audio in different ways. LongCat-Next~\cite{team2026longcat} represents both input and output audio as discrete tokens. UALM~\cite{tian2025ualm} and Audex~\cite{kong2026unified} consume continuous audio encoder representations of input audio but generate discrete codec tokens. Kimi-Audio~\cite{ding2025kimi} combines continuous Whisper~\cite{radford2023robust} features with embeddings of discrete semantic tokens through element-wise addition. Qwen3.5-Omni~\cite{team2026qwen3} and Qwen3-Omni~\cite{xu2025qwen3} instead adopt a Thinker--Talker decomposition: the Thinker processes continuous audio representations for understanding and text generation, while the Talker predicts discrete speech tokens conditioned on features supplied by the Thinker. UniAudio 2.0~\cite{yang2026uniaudio} further decomposes audio into reasoning tokens that emphasize text-aligned semantics and reconstruction tokens that preserve acoustic detail; its ablation results show that using reconstruction tokens alone yields weaker understanding performance than jointly using both token types. In DualSpeechLM~\cite{wang2026dualspeechlm}, the full system adopts semantic inputs and acoustic outputs and achieves a better balance between ASR and TTS performance than baselines that use the same token type for both input and output. Taken together, the findings from UniAudio 2.0 and DualSpeechLM suggest that assigning different representational roles to understanding and generation can help balance their joint performance. Nevertheless, these methods still rely on quantized, codebook-based targets for audio generation, which may introduce information loss; multi-codebook or high-token-rate designs can also incur additional modeling overhead.

Another line of work avoids quantized audio generation targets by directly modeling continuous latents that can be decoded back into waveforms. LatentLM~\cite{sun2024multimodal} introduces next-token diffusion as a general interface for autoregressive modeling of continuous modalities, while DiTAR~\cite{jia2025ditar} and VibeVoice~\cite{peng2025vibevoice} demonstrate the viability of autoregressive continuous-latent modeling for speech synthesis. Ming-Flash-Omni~\cite{ai2025ming} incorporates a continuous acoustic generation pathway into an Omni model alongside multimodal understanding. Other approaches unify understanding and generation around continuous representations in different ways. Ming-UniAudio~\cite{yan2025ming} uses a semantic module to map low-dimensional latents produced by a variational autoencoder (VAE)~\cite{kingma2013auto} into representations consumed by the large language model (LLM), such that understanding and generation share a representation pathway rooted in the same VAE latent space. Audio-Omni~\cite{tian2026audio} uses a frozen multimodal language model to extract semantic representations from text, audio, or video, and conditions a trainable Diffusion Transformer (DiT)~\cite{peebles2023scalable} on these representations for audio generation and editing, whereas UAT~\cite{wang2026uat} combines diffusion over continuous audio latents with masked discrete diffusion for text in a shared dual-stream backbone. A continuous interface alone, however, does not automatically resolve task conflicts. Ablations in Ming-UniAudio show that leaving the shared semantic module unfrozen during early joint training degrades the overall performance of both understanding and generation. The key question is therefore not merely whether continuous representations are used, but whether a shared model can organize task-appropriate continuous input representations for the distinct objectives of understanding and generation.


Motivated by these observations, we propose FireRedAudio, a unified audio-language model with decoupled continuous input representations for understanding and generation. Audio to be transcribed or analyzed is encoded by an Audio Encoder into continuous perceptual representations at $12.5~\mathrm{Hz}$. The speech-generation pathway uses a pretrained deterministic continuous audio autoencoder, termed RedAE (Red Audio Autoencoder), whose latents are aggregated into $6.25~\mathrm{Hz}$ RedAE-Patch representations. Each input audio segment is routed through the pathway appropriate to its role and provided to the shared 9B-parameter LLM together with text instructions. For understanding tasks, the LLM directly generates text tokens. For speech-generation tasks, its hidden states condition a DiT trained with flow-matching~\cite{lipman2022flow} to generate continuous RedAE latents, which are subsequently rendered into $24~\mathrm{kHz}$ waveforms by the RedAE Decoder. The model is jointly optimized with cross-entropy over text tokens and a flow-matching loss for continuous-latent speech generation. FireRedAudio supports ASR and audio understanding, with the latter extending to recordings of up to one hour, as well as zero-shot TTS, Instruct TTS, and speech editing. For structured organization of long-form audio, FireRedAudio demonstrates second-level timestamp accuracy. The model achieves competitive or leading results across a broad evaluation suite.

Our main contributions are summarized as follows:
\begin{itemize}
    \item \textbf{Decoupled continuous audio input representations.} FireRedAudio introduces two input-side continuous representation pathways backed by separate encoders within one trainable autoregressive LLM. To the best of our knowledge, this design has not been reported in prior unified audio-language models. The Audio Encoder pathway serves understanding, whereas the RedAE-Patch pathway serves speech generation; both share the same language backbone without fusing their input representations.
    \item \textbf{Unified capabilities and comprehensive evaluation.} We evaluate FireRedAudio on ASR, audio understanding (MMAU~\cite{sakshi2025mmau} and MMSU~\cite{wang2025mmsu}), zero-shot TTS (Seed-TTS-Eval~\cite{anastassiou2024seed}), Instruct TTS (InstructTTSEval~\cite{huang2025instructttseval}), and speech editing (Ming-Freeform-Audio-Edit~\cite{yan2025ming}). With a 9B-parameter LLM backbone, FireRedAudio achieves competitive or leading results across these tasks.
    \item \textbf{Unified controllable speech generation and editing.} Through one continuous-latent speech-generation pathway, FireRedAudio supports zero-shot TTS, Instruct TTS with natural-language control over vocal timbre, emotion, speaking rate, and volume, and speech editing of both semantic content and acoustic attributes.
    \item \textbf{Structured long-form audio organization with second-level timestamp accuracy.} FireRedAudio organizes recordings of up to one hour into timestamped entries and demonstrates stable time--content alignment under a one-second boundary tolerance on recordings from 5 to 50 minutes.
\end{itemize}

\section{Method}
\label{sec:method}

\subsection{Model Overview}
\label{sec:model_overview}

\begin{figure*}[ht]
    \centering
    \includegraphics[width=0.70\textwidth]{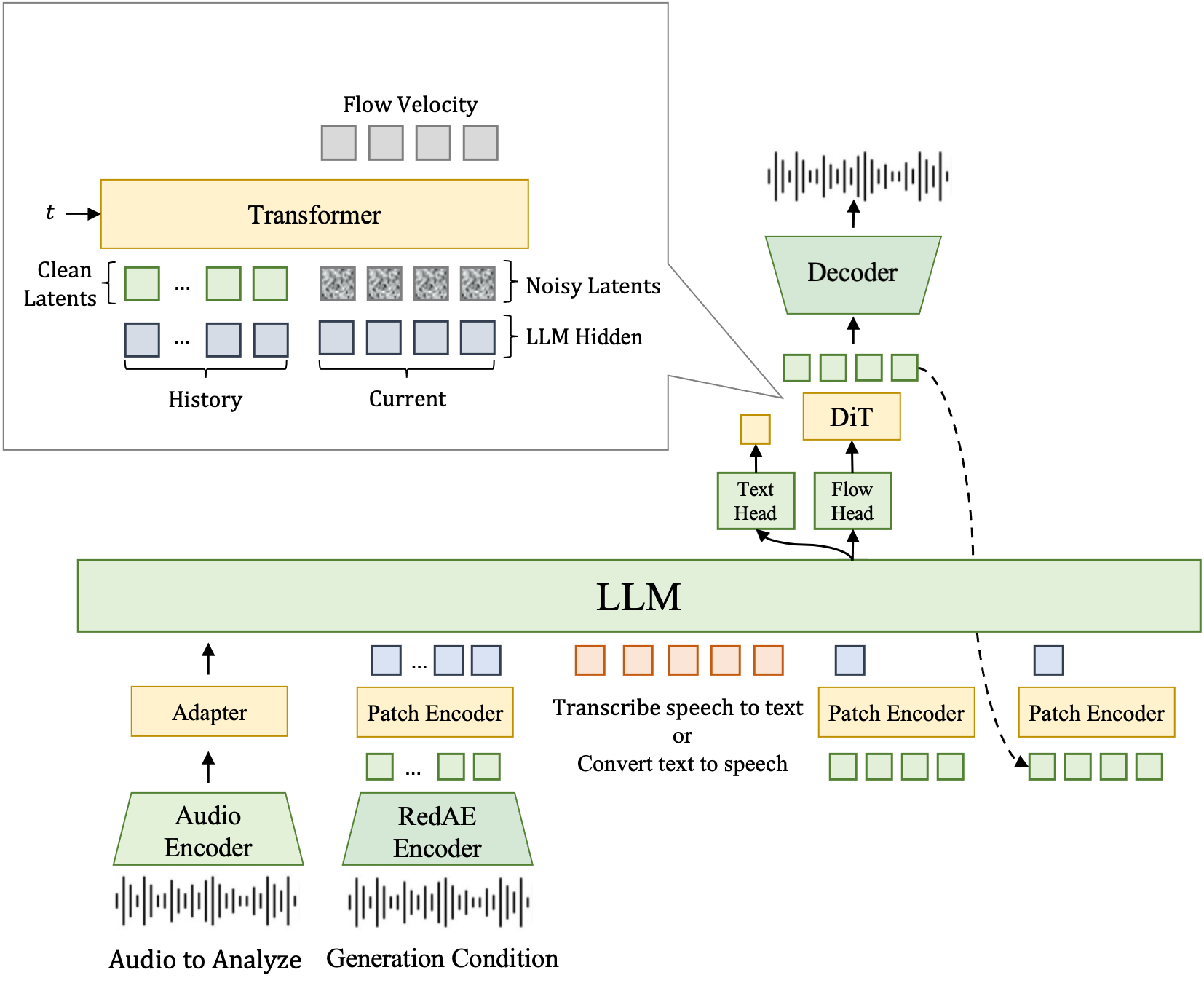}
    \caption{Overview of FireRedAudio. Decoupled continuous pathways encode inputs for understanding and generation. The shared LLM produces text or conditions a DiT to generate RedAE latents, which are decoded into waveforms. The inset illustrates the DiT conditioning for one audio step.}
    \label{fig:fireredaudio_architecture}
\end{figure*}

Understanding and generation both depend on linguistic, paralinguistic, and non-linguistic acoustic information, but require them at different levels of abstraction. Understanding benefits from compact representations that organize task-relevant cues for recognition and analysis, whereas speech generation additionally requires waveform-reconstructible representations that preserve fine-grained acoustic detail. Accordingly, FireRedAudio retains a shared LLM backbone while introducing two decoupled continuous input pathways whose representations are produced by separate encoders and remain unfused. As illustrated in Figure~\ref{fig:fireredaudio_architecture}, the main model comprises an Audio Encoder, an Audio Adapter, a RedAE Encoder, a Patch Encoder, a shared 9B-parameter LLM backbone, and a flow-matching DiT. During inference, a separately loaded RedAE Decoder renders the generated RedAE latents into waveforms.

The input pathway is selected according to the functional role of each audio segment in the current task. Audio to be transcribed or analyzed is processed by the Audio Encoder and Audio Adapter into continuous perceptual representations at $12.5~\mathrm{Hz}$, whereas speech inputs for speech-generation tasks are encoded by the RedAE Encoder into a sequence of RedAE latent frames at $25~\mathrm{Hz}$. We define an \emph{audio step} as four consecutive RedAE latent frames, corresponding to $160~\mathrm{ms}$ of audio. The Patch Encoder aggregates these four frames into one RedAE-Patch representation, yielding an LLM-side frame rate of $6.25~\mathrm{Hz}$. Representations from the two pathways are neither added nor fused; instead, they are inserted at the positions of their corresponding audio segments and provided to the shared LLM together with text instructions. For understanding tasks, the LLM autoregressively generates a textual response. For speech-generation tasks, the LLM provides conditioning to the DiT one audio step at a time, and the DiT generates four RedAE latent frames per step. The newly generated frames are converted into a RedAE-Patch representation by the Patch Encoder and fed back to the LLM as context for the next audio step. Once speech generation terminates, the complete RedAE latent sequence is rendered into a $24~\mathrm{kHz}$ waveform by the RedAE Decoder.

\subsection{Decoupled Continuous Audio Representations}
\label{sec:decoupled_representations}

\subsubsection{Audio Encoder for Understanding}
\label{sec:audio_encoder}

Understanding-oriented representations must preserve linguistic content, paralinguistic cues, and task-relevant acoustic evidence while remaining sufficiently compact to keep the sequence length manageable for long-form audio. The understanding pathway therefore receives a $16~\mathrm{kHz}$ mono waveform and extracts log-Mel features. The Audio Encoder, including its convolutional front end and Transformer, is initialized from the Whisper-large-v3 Encoder\footnote{\url{https://huggingface.co/openai/whisper-large-v3}}~\cite{radford2023robust}, thereby leveraging pretrained multilingual speech representations and acoustic--linguistic alignment priors. A separate Audio Adapter performs additional temporal compression and maps the resulting audio features into the LLM input space.

To control the computational cost of encoding long-form audio, the $100~\mathrm{Hz}$ log-Mel sequence is first partitioned into non-overlapping windows of up to 30 seconds. Within each window, the convolutional front end of the Audio Encoder encodes and downsamples the features to $50~\mathrm{Hz}$, after which the Transformer independently performs local encoding. The window-level outputs are concatenated in their original temporal order and passed to the Audio Adapter, which further downsamples the sequence to $12.5~\mathrm{Hz}$ and projects it into the LLM input embedding space. Long-range interactions across windows are subsequently modeled by the shared LLM. Isolating the additional temporal compression and cross-modal alignment in the Audio Adapter also enables the model to first learn this alignment interface before further optimizing the Audio Encoder.

\subsubsection{RedAE and Patch Encoder for Continuous-Latent Speech Generation}
\label{sec:redae}

\begin{figure*}[ht]
    \centering
    \includegraphics[width=0.75\textwidth]{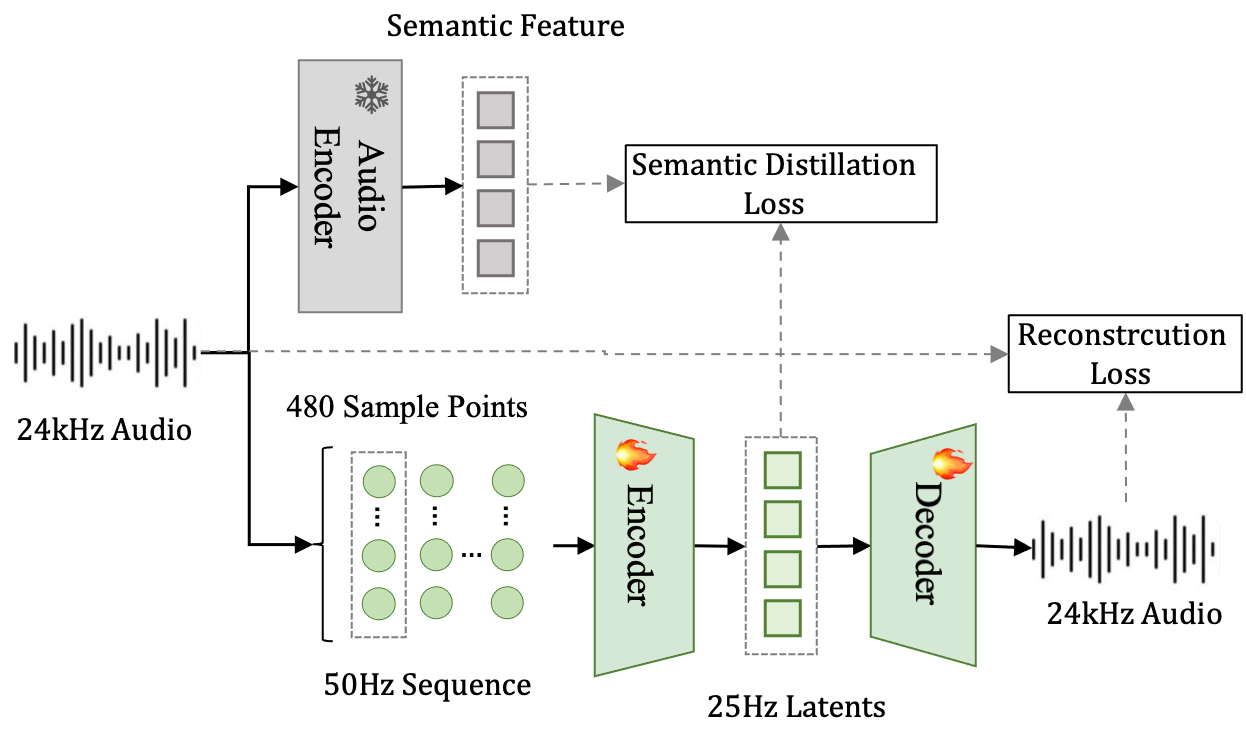}
    \caption{Architecture and pretraining of RedAE. A frozen teacher Audio Encoder provides high-level supervision, while the trainable autoencoder compresses $50~\mathrm{Hz}$ audio frames into $25~\mathrm{Hz}$ latents and reconstructs the waveform through an iSTFT head. Snowflakes and flames denote frozen and trainable modules, respectively.}
    \label{fig:redae_architecture}
\end{figure*}

To obtain waveform-reconstructible continuous representations for speech generation, we independently pretrain RedAE, a deterministic audio autoencoder that maps waveforms directly to continuous latents without variational posterior modeling or Kullback--Leibler (KL) regularization~\cite{joyce2025kullback}. Reconstruction-only training can leave the latent space dominated by low-level acoustic variation~\cite{zhou2026f3}, complicating autoregressive modeling and potentially exacerbating error accumulation. We therefore use a distillation loss to align the RedAE latents with high-level features from a frozen teacher Audio Encoder trained on understanding tasks spanning speech, general audio, and music. This supervision encourages high-level structure in the latents, facilitating their subsequent alignment with the LLM while preserving waveform reconstructibility. Unlike Ming-UniAudio~\cite{yan2025ming}, which appends a separate semantic module after the reconstructible latent representation, RedAE incorporates high-level audio supervision during autoencoder pretraining; the teacher Audio Encoder is discarded afterward and adds no runtime branch.

As shown in Figure~\ref{fig:redae_architecture}, the RedAE Encoder adopts a causal architecture. A $24~\mathrm{kHz}$ waveform is partitioned into non-overlapping blocks of 480 samples to form a $50~\mathrm{Hz}$ frame sequence. Two cascaded Qwen3~\cite{yang2025qwen3} Transformer modules perform contextual encoding and temporal aggregation, respectively. The second module aggregates each pair of adjacent frames, and an output projection produces 64-dimensional RedAE latents at $25~\mathrm{Hz}$. The RedAE Decoder reverses this temporal compression through a linear projection and channel-to-time rearrangement, followed by a causal Qwen3 Transformer and an inverse short-time Fourier transform (iSTFT) head for waveform reconstruction~\cite{siuzdak2024vocos}. RedAE is pretrained on a mixture comprising 50\% clean speech, 25\% noisy speech, 10\% sound effects, and 15\% music. This mixture enables RedAE to learn reconstructible continuous representations across speech, environmental sounds, and music.

Following the X-Codec framework~\cite{ye2025codec}, RedAE is optimized with a hybrid GAN-based objective. Because RedAE does not parameterize a latent posterior, the generator objective omits KL regularization and instead incorporates semantic distillation:

\[
\mathcal{L}_{G}
=\lambda_{\mathrm{rec}}\mathcal{L}_{\mathrm{rec}}
+\lambda_{\mathrm{adv}}\mathcal{L}_{\mathrm{adv}}
+\lambda_{\mathrm{fm}}\mathcal{L}_{\mathrm{fm}}
+\lambda_{\mathrm{distill}}\mathcal{L}_{\mathrm{distill}}.
\]

Here, $\mathcal{L}_{\mathrm{rec}}$, $\mathcal{L}_{\mathrm{adv}}$, and $\mathcal{L}_{\mathrm{fm}}$ denote the reconstruction loss, adversarial loss, and discriminator feature-matching loss, respectively. The term $\mathcal{L}_{\mathrm{distill}}$ is a mean-squared-error (MSE) semantic distillation loss that aligns the RedAE latents with features from the frozen teacher Audio Encoder. This objective is used only for standalone RedAE pretraining. During FireRedAudio training, the pretrained RedAE Encoder is frozen and the RedAE Decoder is not loaded; the Decoder is used only for inference.

Feeding the native RedAE latent frames into the LLM individually would substantially increase the audio sequence length and autoregressive modeling cost. We therefore introduce a Patch Encoder only at the LLM interface. A local Transformer aggregates every four consecutive $25~\mathrm{Hz}$ RedAE latent frames into one RedAE-Patch representation with the same dimensionality as the LLM input embeddings, reducing the LLM-side audio frame rate to $6.25~\mathrm{Hz}$. Waveform reconstruction and DiT-based speech generation remain in the native RedAE latent space.

\subsection{Unified Autoregressive Modeling with a Shared LLM}
\label{sec:unified_llm}

FireRedAudio supports tasks with heterogeneous input--output structures: ASR and audio understanding map audio to text; zero-shot TTS synthesizes target speech from text using reference speech; Instruct TTS follows natural-language instructions to control speech generation; and speech editing transforms a source waveform according to an editing instruction. To unify these tasks within one model, we adopt Qwen3.5-9B\footnote{\url{https://huggingface.co/Qwen/Qwen3.5-9B}} as the shared language backbone and formulate every task as a ChatML conversation. Task-specific system prompts distinguish the tasks, while text instructions, conditioning audio, and target text or speech occupy predefined message roles. After serialization, text tokens use the LLM token embeddings, and each audio segment is routed according to its role through either the Audio Encoder--Adapter pathway or the RedAE--Patch Encoder pathway.

The shared LLM supports two output modes. For ASR and audio understanding, it generates a textual response token by token. For zero-shot TTS, Instruct TTS, and speech editing, the language-model head predicts an audio-start token, a sequence of audio-step tokens, and an audio-end token. The hidden state at each audio step conditions the DiT to generate four RedAE latent frames, which are converted by the Patch Encoder into a RedAE-Patch representation and returned to the LLM as context for the next step. The same interface accommodates different conditions: zero-shot TTS uses reference speech as an audio prefix; Instruct TTS starts from text alone; and speech editing conditions on the source audio and editing instruction, resynthesizing the complete target waveform rather than applying local inpainting to the source RedAE latents.

\subsection{Continuous-Latent Speech Generation with a Flow-Matching DiT}
\label{sec:dit}

For each audio step, the shared LLM produces a hidden state that summarizes the task instruction and all preceding sequence context. The DiT~\cite{peebles2023scalable} uses a local window containing the hidden states of the current and two preceding audio steps to generate the four RedAE latent frames of the current step. This division allows the LLM to model sequence-level context while the DiT handles fine-grained continuous acoustics. We train the DiT with flow-matching~\cite{lipman2022flow} to learn a conditional velocity field from Gaussian noise to the target RedAE latents, avoiding quantized speech tokens.

During training, the frozen RedAE Encoder converts each target waveform into ground-truth RedAE latents, which are grouped into audio steps of four consecutive frames. For each target step, we sample Gaussian noise and a flow-matching time $t$, interpolate between the noise and the four target frames to obtain the noised input, and use the velocity of the corresponding probability path as the regression target. In addition to the noised current-step latents and $t$, the DiT receives two complementary conditioning signals.

The acoustic conditioning comprises eight historical RedAE latent frames from the two preceding audio steps, concatenated with the four noised frames of the current step to form a 12-frame acoustic input. Ground-truth RedAE latents provide the acoustic history during training.
The LLM conditioning comprises the causal hidden state immediately preceding the ground-truth RedAE-Patch of the current step and the corresponding hidden states of the two preceding steps. Because the current state is obtained before the LLM consumes the current ground-truth patch, the target cannot leak into its own condition. The three hidden states are arranged chronologically and repeated four times each to align with the RedAE latent frames.
Missing history in either conditioning window is zero-padded. The LLM conditioning captures the instruction and causal sequence context, whereas the acoustic conditioning promotes local continuity.

The temporally aligned LLM conditioning is concatenated with the 12-frame acoustic input along the feature dimension, while $t$ is injected through a separate time embedding. Only the DiT outputs at the final four positions contribute to the loss; the first eight positions provide acoustic context. Let $\mathcal{P}$ denote the set of valid target audio steps in the training data, and let $\widehat{V}_j$ and $V_j^{\star}$ denote the predicted and target velocities for the $j$-th audio step, respectively. The flow-matching loss is

\[
\mathcal{L}_{\mathrm{flow}}
=\frac{1}{|\mathcal{P}|}
\sum_{j\in\mathcal{P}}
\operatorname{MSE}\!\left(\widehat{V}_j,V_j^{\star}\right).
\]

The MSE is averaged over the four frames and 64 feature dimensions of each target audio step. During training, we randomly drop the LLM conditioning to enable classifier-free guidance~\cite{ho2022classifier} while always retaining the acoustic conditioning.

At inference time, each audio-step token predicted by the language-model head invokes one DiT step. The two-step acoustic-history buffer is initialized with the final two steps of the reference speech for zero-shot TTS and with zeros for the other speech-generation tasks. Source audio for speech editing affects the DiT only through the LLM conditioning and is not inserted into this buffer. Starting from Gaussian noise, the DiT integrates the learned velocity field to generate four RedAE latent frames. These frames update the acoustic-history buffer, which retains the most recent two steps, and are also converted by the Patch Encoder into a RedAE-Patch representation that is returned to the LLM. Generation continues until the language-model head predicts the audio-end token, after which the RedAE Decoder reconstructs a $24~\mathrm{kHz}$ waveform from the complete latent sequence.

\subsection{Joint Training Objective}
\label{sec:joint_objective}

With the RedAE Encoder frozen, FireRedAudio jointly optimizes an autoregressive token objective and a continuous flow-matching objective:

\[
\mathcal{L}
=\lambda_{\mathrm{text}}\mathcal{L}_{\mathrm{text}}
+\lambda_{\mathrm{flow}}\mathcal{L}_{\mathrm{flow}}.
\]

Each loss is normalized over its own valid positions, and we set $\lambda_{\mathrm{text}}=\lambda_{\mathrm{flow}}=1$. Because examples with continuous acoustic targets are sampled less frequently in the mixed training corpus, equal weighting avoids further weakening the continuous acoustic learning signal.

$\mathcal{L}_{\mathrm{text}}$ applies next-token cross-entropy to ordinary text and to the audio-start, audio-step, and audio-end tokens, while excluding prompts, padding, and conditioning audio. Ordinary text and reasoning tokens, together with the audio-start and audio-end tokens, use unit weight. Because audio-step positions are numerous in long speech sequences, we assign them a weight of $w_a=0.01$. These special tokens control only the structure of the output speech sequence; its continuous acoustic content is generated by the DiT.

Let $\mathcal{T}$ and $\mathcal{A}$ denote the unit-weight and audio-step positions, respectively, and let $\ell_i=-\log p(y_i\mid y_{<i},\mathrm{context})$. We compute

\[
\mathcal{L}_{\mathrm{text}}
=
\frac{
\sum_{i\in\mathcal{T}}\ell_i
+w_a\sum_{i\in\mathcal{A}}\ell_i
}{
|\mathcal{T}|+w_a|\mathcal{A}|
},
\qquad w_a=0.01.
\]

Thus, $w_a$ weights both the numerator and the valid-token denominator rather than scaling the final loss as a whole. The denominator is aggregated across data-parallel workers to obtain a global weighted-token average. The joint objective updates the shared LLM, Audio Encoder, Audio Adapter, Patch Encoder, and DiT.

\section{Training Strategy}
\label{sec:training_strategy}

FireRedAudio combines pretrained modules with newly initialized cross-modal and speech-generation components. The Audio Encoder, LLM, and RedAE are pretrained, whereas the Audio Adapter must establish the understanding interface and the Patch Encoder and DiT must establish the continuous-latent speech-generation interface with the shared LLM. Jointly optimizing all trainable components from the outset would expose the LLM to unstable understanding and speech-generation interfaces. We therefore adopt a five-stage progressive strategy. Adapter alignment first stabilizes the understanding pathway, after which Audio Encoder adaptation broadens it. Unified mid-training jointly optimizes all trainable modules and introduces the continuous-latent speech-generation objective. Multitask post-training adds Instruct TTS and speech editing while strengthening fine-grained, structured, and multi-step audio understanding. Finally, long-context extension expands audio understanding to recordings of up to one hour. The RedAE Encoder remains frozen throughout main-model training.

\subsection{Data Mixture and Sampling}
\label{sec:data_mixture}

The data required for unified multitask training vary greatly in scale. Text-only language-modeling, ASR, and zero-shot TTS data are substantially more abundant than data for Instruct TTS, speech editing, and fine-grained audio understanding; further imbalances exist within each task family across languages, acoustic domains, and data sources. Sampling in proportion to raw data scale would therefore allow large tasks and datasets to dominate optimization. Conneau et al.~\cite{conneau2020unsupervised} apply power-law smoothing to a sampling distribution over languages in multilingual speech pretraining. We extend this idea to a two-level hierarchy: we first sample a task family and then a constituent dataset.

Let $W_i$ denote the configured sampling weight of the $i$-th task family and $w_{ij}$ the weight of its $j$-th constituent dataset. The two-level sampling probabilities are

\[
p(i)
=
\frac{W_i^{\alpha_{\mathrm{task}}}}
{\sum_k W_k^{\alpha_{\mathrm{task}}}},
\qquad
p(j\mid i)
=
\frac{w_{ij}^{\alpha_{\mathrm{data}}}}
{\sum_l w_{il}^{\alpha_{\mathrm{data}}}}.
\]

We set the task-level and dataset-level smoothing exponents to $\alpha_{\mathrm{task}}=0.5$ and $\alpha_{\mathrm{data}}=0.7$, respectively. The smaller task-level exponent provides stronger balancing across task families, whereas the dataset-level exponent moderates scale differences among datasets within the same family. We adjust task-family weights across training stages according to their respective objectives, thereby controlling how frequently each task is sampled during optimization.

We use the term multimodal token to refer collectively to a text token or a continuous audio representation entering the LLM. The training volume of each stage is measured by the total number of non-padding multimodal tokens processed during optimization and reported in billions of tokens (B tokens). Table~\ref{tab:training_stages} summarizes the data focus, training volume, and maximum per-example sequence length of each stage.

\begin{table}[ht]
\centering
\small
\setlength{\tabcolsep}{4pt}
\renewcommand{\arraystretch}{1.15}
\caption{Overview of the five-stage training strategy. Volumes are measured in billions of non-padding multimodal tokens and rounded to the nearest billion. Sequence length denotes the maximum effective length per example.}
\label{tab:training_stages}
\begin{tabularx}{\textwidth}{
    @{}
    l
    >{\raggedright\arraybackslash}X
    c
    c
    @{}
}
\toprule
\textbf{Stage}
& \textbf{Data Focus}
& \textbf{Volume}
& \textbf{Sequence Length} \\
\midrule
Adapter Alignment
& Chinese- and English-language ASR; multilingual ASR
& $180$B
& $8\mathrm{k}$ \\
Audio Encoder Adaptation
& ASR and understanding of speech, general audio, and music
& $390$B
& $8\mathrm{k}$ \\
Unified Mid-training
& Text, ASR, audio understanding, zero-shot TTS, and audio--text interleaving
& $990$B
& $8\mathrm{k}$ \\
Multitask Post-training
& Instruct TTS and speech editing; retained ASR, audio understanding, text-only instruction data, and zero-shot TTS
& $511$B
& $8\mathrm{k}$ \\
Long-context Extension
& Long-form audio understanding for recordings of up to one hour; retained regular-length multitask data
& $591$B
& $200\mathrm{k}$ \\
\bottomrule
\end{tabularx}
\end{table}

\subsection{Progressive Training}
\label{sec:progressive_training}

\subsubsection{Adapter Alignment}
\label{sec:adapter_alignment}

Although the Audio Encoder provides pretrained speech representations and the LLM possesses pretrained language-modeling capabilities, the randomly initialized Audio Adapter cannot yet connect them reliably. Updating both pretrained components before this interface is aligned could allow noise from the cross-modal mapping to disrupt their existing representations. We therefore use $80$B Chinese- and English-language ASR tokens and $100$B multilingual ASR tokens, as transcription provides dense and well-defined audio--text supervision. The Audio Encoder and LLM are frozen, and only the Audio Adapter is optimized with autoregressive text cross-entropy. Because the Adapter is randomly initialized and trained in isolation at this stage, its learning rate is linearly warmed up for 500 steps to $2\times10^{-4}$ and then held constant. This stage teaches the Audio Adapter to temporally compress the Audio Encoder features and map them into the LLM input space, establishing a stable interface before the audio front end is further optimized.

\subsubsection{Audio Encoder Adaptation}
\label{sec:audio_encoder_adaptation}

Because the transcription supervision used in the first stage primarily captures linguistic content, the second stage broadens the training signal to cover paralinguistic and speaker information, environmental sounds, music, and temporal relations. In addition to $120$B Chinese- and English-language ASR tokens and $120$B multilingual ASR tokens, this stage uses $150$B audio-understanding tokens to expand the perceptual scope beyond transcription. Once the cross-modal interface has stabilized, we unfreeze the Audio Encoder and jointly optimize it with the Audio Adapter using autoregressive text cross-entropy, while keeping the LLM frozen as a fixed language backbone. We retain a peak learning rate of $2\times10^{-4}$ to adapt the audio front end to the broader perceptual supervision, using a $2.5\mathrm{k}$-step warm-up followed by linear decay. This setup adapts the transcription-oriented Audio Encoder into a general audio-perception module.

\subsubsection{Unified Mid-training}
\label{sec:unified_mid_training}

Having aligned the understanding pathway, we expand the curriculum to language modeling, audio understanding, and zero-shot TTS. At this stage, we jointly optimize the Audio Encoder, Audio Adapter, LLM, Patch Encoder, and DiT for the first time. The mixture comprises $990$B multimodal tokens: $160$B Chinese- and English-language ASR tokens, $200$B multilingual ASR tokens, $250$B audio-understanding tokens, $200$B text-only tokens, $120$B zero-shot TTS tokens, and $60$B audio--text interleaved tokens. The text-only data preserve language modeling and instruction following. In interleaved examples, preceding audio is encoded by the understanding pathway as LLM context and only the subsequent text is predicted, strengthening audio--language alignment. The zero-shot TTS data introduce the continuous-latent speech-generation objective.

We do not separately pretrain the Patch Encoder or DiT. The Patch Encoder learns to map RedAE latents into RedAE-Patch representations that the shared LLM can consume as audio context, while the DiT learns to generate continuous RedAE latents conditioned on the corresponding LLM hidden states. They are therefore trained jointly through the complete continuous-latent speech-generation loop. With the RedAE Encoder frozen to maintain a stationary latent space, all trainable modules are jointly optimized using the weighted text cross-entropy and flow-matching losses defined in Section~\ref{sec:joint_objective}. The former supervises text outputs and audio-sequence control tokens, whereas the latter supervises continuous acoustic content. Because the pretrained LLM is shared across understanding and generation, we update it conservatively. Meanwhile, the Audio Encoder and Audio Adapter continue adapting to the unified training distribution, while the Patch Encoder and DiT establish the continuous-latent speech-generation loop. Accordingly, we use a smaller peak learning rate for the LLM than for the audio modules. Specifically, we linearly warm up the learning rates for $2.5\mathrm{k}$ steps to peak values of $3\times10^{-5}$ for the LLM backbone and $2\times10^{-4}$ for the four audio modules, followed by linear decay. The goal of this stage is to establish a shared foundation for language modeling, audio perception, and continuous-latent speech generation rather than to specialize the model for individual tasks.

\subsubsection{Multitask Post-training}
\label{sec:multitask_post_training}

Whereas mid-training establishes a broad foundation from large-scale data, the fourth stage emphasizes instruction following. To preserve previously acquired capabilities, this stage uses $13$B high-quality Chinese- and English-language ASR tokens, $4$B multilingual ASR tokens, $300$B audio-understanding tokens, $60$B text-only instruction tokens, and $100$B zero-shot TTS tokens. It additionally introduces $22$B Instruct TTS tokens and $12$B speech-editing tokens, yielding approximately $511$B tokens. Explicit chain-of-thought (CoT) supervision is added to Instruct TTS and speech-editing data, as well as to audio-understanding examples that require multi-step reasoning. This supervision encourages the model to reason over synthesis conditions, editing instructions, or audio evidence before producing the target speech or textual answer.

Post-training replaces non-instructional text-only data with text-only instruction data, while retaining ASR, zero-shot TTS, and audio-understanding supervision to mitigate forgetting. The set of trainable modules and the joint objective remain unchanged from mid-training. Because Instruct TTS and speech editing require the shared LLM to learn new instruction-conditioned output patterns, we retain a peak learning rate of $3\times10^{-5}$ for the LLM backbone. By this stage, both the Audio Encoder--Adapter understanding pathway and the Patch Encoder--DiT continuous-latent speech-generation pathway have been established. We therefore reduce the peak learning rate of all four audio modules to $1\times10^{-4}$ for task-specific refinement. After a 500-step warm-up, the two learning-rate levels decay linearly.

\subsubsection{Long-context Extension}
\label{sec:long_context_extension}

Before long-context extension, audio-understanding examples are limited to at most 5 minutes, which is insufficient for integrating evidence across distant temporal regions in long-form recordings. In the final stage, we increase the maximum per-example effective length of the unified text--audio sequence from $8\mathrm{k}$ to $200\mathrm{k}$ and extend the maximum audio duration for understanding to one hour. With this extended context, FireRedAudio is trained for global summarization, temporal and content localization, cross-segment analysis, complex question answering, and structured organization of long-form audio.

To mitigate forgetting during long-context specialization, we retain the complete post-training mixture and add $80$B long-form audio-understanding tokens, yielding approximately $591$B tokens in total. The trainable modules and joint objective remain unchanged from post-training. Because this stage extends context length rather than introducing a new modality interface, we further reduce the peak learning rates to $1\times10^{-5}$ for the LLM backbone and $3\times10^{-5}$ for the Audio Encoder, Audio Adapter, Patch Encoder, and DiT. After a 500-step warm-up, the two learning-rate levels decay linearly. 

\subsection{Optimization Configuration}
\label{sec:optimization_configuration}

Table~\ref{tab:training_hyperparameters} reports the stage-wise learning rates of the LLM backbone, Audio Encoder, Audio Adapter, Patch Encoder, and DiT, together with the learning-rate schedules, frozen modules, and per-replica packing budgets. To reduce padding for variable-length multimodal examples, we construct each packed microbatch under stage-specific limits on the multimodal-token count and separate audio-duration budgets for the understanding and speech-generation pathways. These constraints keep per-replica memory use stable across heterogeneous multimodal batches.

\begin{table}[ht]
\centering
\small
\setlength{\tabcolsep}{3pt}
\renewcommand{\arraystretch}{1.18}
\renewcommand{\tabularxcolumn}[1]{m{#1}}
\caption{Optimization configurations for the five training stages. A dash indicates that the corresponding module is not optimized in that stage.}
\label{tab:training_hyperparameters}
\resizebox{\textwidth}{!}{%
\begin{tabularx}{\textwidth}{
    @{}
    >{\raggedright\arraybackslash}m{0.26\textwidth}
    *{5}{>{\centering\arraybackslash}X}
    @{}
}
\toprule
\multicolumn{1}{c}{\multirow{2}{*}{\textbf{Hyperparameter}}}
& \multicolumn{2}{c}{\shortstack[c]{\textbf{Understanding}\\\textbf{Alignment}}}
& \multicolumn{2}{c}{\shortstack[c]{\textbf{Unified Multitask}\\\textbf{Training}}}
& \multicolumn{1}{c}{\shortstack[c]{\textbf{Long-context}\\\textbf{Extension}}} \\
\cmidrule(lr){2-3}\cmidrule(lr){4-5}\cmidrule(l){6-6}
& \shortstack[c]{\textbf{Adapter}\\\textbf{Alignment}}
& \shortstack[c]{\textbf{Encoder}\\\textbf{Adaptation}}
& \shortstack[c]{\textbf{Unified}\\\textbf{Mid-training}}
& \shortstack[c]{\textbf{Multitask}\\\textbf{Post-training}}
& \shortstack[c]{\textbf{Extension}} \\
\midrule
Peak LR (LLM backbone)
& frozen
& frozen
& $3{\times}10^{-5}$
& $3{\times}10^{-5}$
& $1{\times}10^{-5}$ \\
Peak LR (Audio Encoder)
& frozen
& $2{\times}10^{-4}$
& $2{\times}10^{-4}$
& $1{\times}10^{-4}$
& $3{\times}10^{-5}$ \\
Peak LR (Audio Adapter)
& $2{\times}10^{-4}$
& $2{\times}10^{-4}$
& $2{\times}10^{-4}$
& $1{\times}10^{-4}$
& $3{\times}10^{-5}$ \\
Peak LR (Patch Encoder and DiT)
& --
& --
& $2{\times}10^{-4}$
& $1{\times}10^{-4}$
& $3{\times}10^{-5}$ \\
\addlinespace[2pt]
LR schedule
& constant
& linear
& linear
& linear
& linear \\
Warm-up steps
& $500$
& $2.5\mathrm{k}$
& $2.5\mathrm{k}$
& $500$
& $500$ \\
\addlinespace[2pt]
Max packed tokens per replica
& $36\mathrm{k}$
& $36\mathrm{k}$
& $36\mathrm{k}$
& $36\mathrm{k}$
& $200\mathrm{k}$ \\
Max understanding audio duration per replica (s)
& $1{,}500$
& $1{,}500$
& $1{,}000$
& $1{,}000$
& $3{,}600$ \\
Max speech-generation duration per replica (s)
& --
& --
& $610$
& $610$
& $610$ \\
\bottomrule
\end{tabularx}%
}
\end{table}

\section{Experiments}
\label{sec:experiments}

\subsection{ASR and Audio Understanding}
\label{sec:asr_audio_understanding}

\begin{table}[ht]
\centering
\small
\setlength{\tabcolsep}{8pt}
\renewcommand{\arraystretch}{1.10}
\caption{Audio-understanding accuracy (\%) on MMAU and MMSU. Higher is better. Best results are shown in bold. Results marked with $^{*}$ are obtained from our own evaluation.}
\label{tab:audio_understanding}
\begin{tabular}{@{}lccc@{}}
\toprule
\textbf{Model}
& \textbf{MMAU} {\scriptsize\textbf{test-mini}}
& \textbf{MMAU} {\scriptsize\textbf{test}}
& \textbf{MMSU} \\
\midrule
Step-Audio-R1.1~\cite{tian2025step}             & 77.7 & -    & 75.9 \\
Step-Audio 2~\cite{wu2025step}                   & 78.0 & -    & -   \\
MiMo-Audio-7B-Instruct~\cite{zhang2025mimo}      & 74.9 & -    & 61.7 \\
Kimi-Audio~\cite{ding2025kimi}                   & 65.2 & -    & -   \\
LongCat-Next~\cite{team2026longcat}              & 76.4 & -    & -   \\
Qwen3-Omni-30B-A3B-Instruct~\cite{xu2025qwen3}   & 77.5 & -    & 69.0 \\
Gemini 3.1 Pro~\cite{gemini31_pro}                & 80.7$^{*}$ & 78.8$^{*}$ & 82.7$^{*}$ \\
Qwen3.5-Omni-Plus~\cite{team2026qwen3}            & 81.4$^{*}$ & 79.9$^{*}$ & 80.7$^{*}$ \\
\textbf{FireRedAudio}                & \textbf{82.0} & \textbf{80.9}  & \textbf{83.3} \\
\bottomrule
\end{tabular}
\end{table}

We evaluate audio understanding on MMAU and MMSU. MMAU~\cite{sakshi2025mmau} contains 10,000 multiple-choice questions covering information extraction and reasoning over speech, environmental sounds, and music; we report its 1,000-question \texttt{test-mini} and 9,000-question \texttt{test} splits. MMSU~\cite{wang2025mmsu} contains 5,000 questions across 47 perception and reasoning tasks and evaluates fine-grained English speech information, including linguistic content, phonology, speaker characteristics, and speaking style. Together, the two benchmarks assess both general audio-understanding breadth and detailed spoken-language understanding. Both report accuracy (ACC, \%), where higher is better.

As shown in Table~\ref{tab:audio_understanding}, FireRedAudio achieves the highest accuracy in all three settings. The consistent advantage across MMAU and MMSU shows strong coverage of both broad audio understanding and fine-grained spoken-language understanding. These results further suggest that the dedicated Audio Encoder pathway preserves strong understanding performance without forcing its representations to accommodate the reconstruction-oriented requirements of speech generation.

Our ASR suite spans Mandarin, English, multilingual, dialectal, web, meeting, and singing speech. It includes AISHELL-1~\cite{bu2017aishell}, AISHELL-2 \texttt{test-ios}~\cite{du2018aishell}, WenetSpeech \texttt{Test\_Net}/\texttt{Test\_Meeting}~\cite{zhang2022wenetspeech}, LibriSpeech \texttt{test-clean}/\texttt{test-other}~\cite{panayotov2015librispeech}, KeSpeech~\cite{tang2021kespeech}, and Opencpop~\cite{wang2022opencpop}. For FLEURS~\cite{fleurs2022arxiv}, we report English, Chinese, and the macro-average over all 102 languages. CER or WER is used as appropriate, and lower is better.

Table~\ref{tab:asr_results} shows that FireRedAudio achieves the lowest reported error rates on LibriSpeech \texttt{test-clean}, FLEURS English, and the FLEURS-102 macro-average, reaching 14.94\% on the latter. It also ranks second on WenetSpeech \texttt{Test\_Meeting} and among the top three reported systems on AISHELL-1, KeSpeech, and Opencpop. Overall, these results demonstrate broad ASR capability across languages and acoustic conditions within a unified model that also supports audio understanding and speech generation.

\begin{table}[ht]
\centering
\scriptsize
\setlength{\tabcolsep}{2.2pt}
\renewcommand{\arraystretch}{1.10}
\caption{ASR results across speech, multilingual, and singing-voice benchmarks. CER or WER is reported as appropriate, and lower is better. Results marked with $^{*}$ are obtained from our own evaluation.}
\label{tab:asr_results}
\resizebox{\textwidth}{!}{%
\begin{tabular}{@{}lcccccccc@{}}
\toprule
\textbf{Model}
& \begin{tabular}[c]{@{}c@{}}\textbf{AISHELL-1}\end{tabular}
& \begin{tabular}[c]{@{}c@{}}\textbf{AISHELL-2}\\[-1pt]{\tiny\textbf{test-ios}}\end{tabular}
& \begin{tabular}[c]{@{}c@{}}\textbf{WenetSpeech}\\[-1pt]{\tiny\textbf{Net | Meeting}}\end{tabular}
& \begin{tabular}[c]{@{}c@{}}\textbf{LibriSpeech}\\[-1pt]{\tiny\textbf{clean | other}}\end{tabular}
& \begin{tabular}[c]{@{}c@{}}\textbf{FLEURS}\\[-1pt]{\tiny\textbf{en | zh}}\end{tabular}
& \begin{tabular}[c]{@{}c@{}}\textbf{FLEURS-102}\footnotemark\\[-1pt]{\tiny\textbf{average}}\end{tabular}
& \begin{tabular}[c]{@{}c@{}}\textbf{KeSpeech}\end{tabular}
& \begin{tabular}[c]{@{}c@{}}\textbf{Opencpop}\end{tabular} \\
\midrule
Step-Audio 2~\cite{wu2025step}                   & 0.63  & 2.10  & 4.67~|~4.75  & 1.17~|~2.42 & 3.03~|~2.68 & -     & 3.63  & -   \\
MiMo-Audio-7B-Instruct~\cite{zhang2025mimo}      & 1.65  & -    & -            & 3.50~|~-    & -          & -     & -    & -   \\
Ming-UniAudio-16B-A3B~\cite{yan2025ming}         & -    & 2.84  & -            & 1.62~|~-    & -          & -     & -    & -   \\
Kimi-Audio~\cite{ding2025kimi}                   & 0.60  & 2.56  & 5.37~|~6.28  & 1.28~|~2.42 & 4.44~|~2.69 & -     & -    & -   \\
LongCat-Next~\cite{team2026longcat}              & 1.47  & 2.82  & 5.98~|~8.19  & 1.63~|~3.42 & 5.24~|~3.24 & -     & -    & -   \\
Qwen3-Omni-30B-A3B-Instruct~\cite{xu2025qwen3}   & -    & -    & 4.69~|~5.89  & 1.22~|~2.48 & 2.72~|~2.20 & -     & -    & 1.54 \\
Gemini 3.1 Pro~\cite{gemini31_pro}                & 3.66$^{*}$ & 7.10$^{*}$ & 11.53~|~14.21 & 3.36~|~4.41 & 2.97$^{*}$~|~4.28$^{*}$ & 18.23$^{*}$ & 23.67 & 6.83 \\
Qwen3.5-Omni-Plus~\cite{team2026qwen3}            & 0.82$^{*}$ & 2.26$^{*}$ & 4.30~|~5.84  & 1.11~|~2.23 & 3.33$^{*}$~|~2.46$^{*}$ & 23.66$^{*}$ & 3.46  & 1.49 \\
\textbf{FireRedAudio}                             & 0.71  & 2.63  & 5.18~|~5.33  & 0.67~|~2.91 & 2.53~|~3.14 & 14.94  & 4.82  & 1.63 \\
\bottomrule
\end{tabular}%
}
\end{table}
\footnotetext{Detailed per-language FLEURS-102 results are provided in~\ref{app:multilingual_asr_results}.}

\subsection{Zero-Shot TTS}
\label{sec:tts_evaluation}

\begin{table}[ht]
\centering
\small
\setlength{\tabcolsep}{3.2pt}
\renewcommand{\arraystretch}{1.12}
\caption{Zero-shot TTS results on Seed-TTS-Eval. Lower CER and WER and higher SIM indicate better performance. The average columns report the mean of the Chinese and English scores. The best and second-best results in each column are shown in bold and underlined, respectively.}
\label{tab:seed_tts_eval}
\begin{tabular}{@{}lcccccc@{}}
\toprule
\multirow[c]{2}{*}[-3pt]{\strut\textbf{Model}}
& \multicolumn{2}{c}{\textbf{Seed-ZH}}
& \multicolumn{2}{c}{\textbf{Seed-EN}}
& \multicolumn{2}{c}{\textbf{Average}} \\
\cmidrule(lr){2-3}\cmidrule(lr){4-5}\cmidrule(lr){6-7}
& \textbf{CER (\%) $\downarrow$}
& \textbf{SIM $\uparrow$}
& \textbf{WER (\%) $\downarrow$}
& \textbf{SIM $\uparrow$}
& \textbf{CER/WER (\%) $\downarrow$}
& \textbf{SIM $\uparrow$} \\
\midrule
\multicolumn{7}{@{}l}{\textit{Specialized TTS models}} \\
Seed-TTS~\cite{anastassiou2024seed} & 1.12 & \textbf{0.80} & 2.25 & \textbf{0.76} & 1.69 & \textbf{0.78} \\
FireRedTTS~\cite{guo2024fireredtts} & 1.51 & 0.65 & 3.82 & 0.53 & 2.67 & 0.59 \\
FireRedTTS-2~\cite{xie2025fireredtts} & 1.14 & 0.74 & 1.95 & 0.65 & 1.55 & 0.69 \\
DiTAR-1B~\cite{jia2025ditar} & 1.02 & 0.75 & 1.69 & \underline{0.74} & 1.36 & \underline{0.75} \\
F5-TTS~\cite{chen2025f5} & 1.56 & 0.74 & 1.83 & 0.65 & 1.70 & 0.70 \\
CosyVoice 2~\cite{du2024cosyvoice} & 1.45 & 0.75 & 2.57 & 0.65 & 2.01 & 0.70 \\
CosyVoice 3-1.5B~\cite{du2025cosyvoice} & 1.12 & \underline{0.78} & 2.21 & 0.72 & 1.67 & \underline{0.75} \\
\addlinespace[2pt]
\multicolumn{7}{@{}l}{\textit{Unified audio models}} \\
MiMo-Audio-7B-Instruct~\cite{zhang2025mimo} & 1.96 & -- & 5.37 & -- & 3.67 & -- \\
Qwen2.5-Omni-7B (RL)~\cite{Qwen2.5-Omni} & 1.42 & 0.75 & 2.33 & 0.64 & 1.88 & 0.70 \\
Qwen3-Omni-30B-A3B-Instruct~\cite{xu2025qwen3} & 1.07 & -- & \textbf{1.39} & -- & \underline{1.23} & -- \\
Ming-UniAudio-16B-A3B~\cite{yan2025ming} & \underline{0.95} & 0.70 & 1.85 & 0.58 & 1.40 & 0.64 \\
\textbf{FireRedAudio} & \textbf{0.83} & 0.74 & \underline{1.56} & 0.68 & \textbf{1.20} & 0.71 \\
\bottomrule
\end{tabular}
\end{table}

We evaluate zero-shot TTS on Seed-TTS-Eval~\cite{anastassiou2024seed}, which contains 1,000 English samples from Common Voice~\cite{ardila2020common} and 2,000 Chinese samples from DiDiSpeech-2~\cite{guo2021didispeech}. Each example pairs reference speech with target text. For FireRedAudio, we compute Chinese CER with Paraformer-zh\footnote{\url{https://huggingface.co/funasr/paraformer-zh}}~\cite{gao2022paraformer}, English WER with Whisper-large-v3~\cite{radford2023robust}, and speaker similarity (SIM) with WavLM-large\footnote{\url{https://huggingface.co/microsoft/wavlm-large}}~\cite{chen2022wavlm}. Lower CER/WER and higher SIM are better; the average columns summarize the Chinese and English scores.

As reported in Table~\ref{tab:seed_tts_eval}, FireRedAudio achieves the best content accuracy on Seed-ZH and ranks second on Seed-EN, yielding the lowest average content-error score of 1.20\% among all compared systems. Its average SIM is 0.71, the highest among unified models that report speaker similarity in both languages. Although its speaker similarity remains below the strongest specialized TTS systems, FireRedAudio uses no dedicated speaker embedding and instead conditions synthesis directly on the RedAE representation of the reference speech. These results suggest that jointly supporting audio understanding and speech generation does not prevent accurate synthesis of the target text or preservation of reference-speaker cues. The strong content accuracy is also consistent with the architectural division of labor: the shared LLM models the target text and generation context, while the DiT--RedAE pathway generates the corresponding continuous acoustics.

\subsection{Instruct TTS}
\label{sec:instruct_tts_evaluation}

\begin{table}[ht]
\centering
\small
\setlength{\tabcolsep}{4pt}
\renewcommand{\arraystretch}{1.12}
\caption{Instruction-following accuracy (\%) on InstructTTSEval. APS, DSD, and RP evaluate structured attribute control, adherence to free-form style descriptions, and role/scenario inference, respectively. VD denotes Voice Design. Higher values indicate better performance on all metrics. Best results are shown in bold. All results are from our reevaluation using Gemini 2.5 Pro.}
\label{tab:instruct_tts_eval}
\begin{tabular}{@{}lcccccc@{}}
\toprule
\multirow[c]{2}{*}[-3pt]{\strut\textbf{Model}}
& \multicolumn{3}{c}{\textbf{ZH}}
& \multicolumn{3}{c}{\textbf{EN}} \\
\cmidrule(lr){2-4}\cmidrule(lr){5-7}
& \textbf{APS $\uparrow$}
& \textbf{DSD $\uparrow$}
& \textbf{RP $\uparrow$}
& \textbf{APS $\uparrow$}
& \textbf{DSD $\uparrow$}
& \textbf{RP $\uparrow$} \\
\midrule
VoiceSculptor-VD~\cite{hu2026voicesculptor} & 74.6 & 63.5 & 62.0 & -- & -- & -- \\
MOSS-VoiceGenerator~\cite{huang2026moss} & 71.6 & 72.5 & 61.3 & 58.8 & 71.8 & 61.6 \\
Ming-Omni-TTS-16B-A3B~\cite{mingomnitts} & 84.6 & 70.7 & 56.0 & -- & -- & -- \\
Qwen3-TTS-VD~\cite{hu2026qwen3} & 83.7 & 81.7 & 65.8 & 76.4 & 81.4 & 64.2 \\
FireRedAudio & \textbf{86.0} & \textbf{84.1} & \textbf{70.1} & \textbf{81.1} & \textbf{83.6} & \textbf{70.3} \\
\bottomrule
\end{tabular}
\end{table}

We evaluate Instruct TTS on InstructTTSEval~\cite{huang2025instructttseval}, which contains 6,000 examples in total across Chinese and English. Its three tasks increase in abstraction: Acoustic-Parameter Specification (APS) gives structured vocal attributes, Descriptive-Style Directive (DSD) uses free-form style descriptions, and Role-Play (RP) requires suitable vocal characteristics to be inferred from a role or scenario. Because the benchmark's original Gemini 2.5 Pro Preview evaluator is no longer available, we reevaluate all systems with Gemini 2.5 Pro~\cite{gemini25_pro} using the official prompt. We report instruction-following accuracy, where higher is better.

As shown in Table~\ref{tab:instruct_tts_eval}, FireRedAudio achieves the highest accuracy in all six Chinese and English settings. The gains span structured attribute control, free-form style descriptions, and role/scenario inference, demonstrating consistent instruction following across different levels of abstraction. The largest margin is 6.1 percentage points on EN RP, suggesting particular strength in inferring unstated speaking styles from contextual cues. This behavior may benefit from the explicit CoT supervision used during post-training, which encourages the shared LLM to infer an appropriate speaking style before conditioning the DiT acoustic generator.

\subsection{Speech Editing}
\label{sec:speech_editing_evaluation}

\begin{table}[ht]
\centering
\small
\setlength{\tabcolsep}{5pt}
\renewcommand{\arraystretch}{1.10}
\caption{Semantic-editing results on Ming-Freeform-Audio-Edit. Lower WER/no-edit WER and higher SIM/ACC indicate better performance. Best results are shown in bold.}
\label{tab:semantic_speech_editing}
\begin{tabular}{@{}lllcc@{}}
\toprule
\begin{tabular}[c]{@{}l@{}}\textbf{Task}\end{tabular}
& \begin{tabular}[c]{@{}l@{}}\textbf{Setting}\end{tabular}
& \begin{tabular}[c]{@{}l@{}}\textbf{Metric}\end{tabular}
& \begin{tabular}[c]{@{}c@{}}\textbf{Ming-UniAudio-Edit}~\cite{yan2025ming}\\\textbf{(ZH | EN)}\end{tabular}
& \begin{tabular}[c]{@{}c@{}}\textbf{FireRedAudio}\\\textbf{(ZH | EN)}\end{tabular} \\
\midrule
\multirow{8}{*}{Deletion}
& \multirow{4}{*}{basic}
& WER (\%) $\downarrow$ & 11.89~|~14.85 & \textbf{10.82}~|~\textbf{12.78} \\
& & SIM $\uparrow$ & \textbf{0.78}~|~0.76 & \textbf{0.78}~|~\textbf{0.79} \\
& & ACC (\%) $\uparrow$ & \textbf{100.00}~|~82.22 & \textbf{100.00}~|~\textbf{97.78} \\
& & no-edit WER (\%) $\downarrow$ & 11.49~|~24.26 & \textbf{10.70}~|~\textbf{23.16} \\
\cmidrule(lr){2-5}
& \multirow{4}{*}{open}
& WER (\%) $\downarrow$ & 22.92~|~27.60 & \textbf{10.49}~|~\textbf{16.65} \\
& & SIM $\uparrow$ & \textbf{0.81}~|~0.74 & 0.80~|~\textbf{0.80} \\
& & ACC (\%) $\uparrow$ & 82.92~|~85.00 & \textbf{89.32}~|~\textbf{86.50} \\
& & no-edit WER (\%) $\downarrow$ & 17.50~|~35.21 & \textbf{7.84}~|~\textbf{25.43} \\
\midrule
\multirow{8}{*}{Insertion}
& \multirow{4}{*}{basic}
& WER (\%) $\downarrow$ & 3.42~|~6.63 & \textbf{3.28}~|~\textbf{4.98} \\
& & SIM $\uparrow$ & \textbf{0.83}~|~0.79 & \textbf{0.83}~|~\textbf{0.84} \\
& & ACC (\%) $\uparrow$ & 80.00~|~71.43 & \textbf{83.53}~|~\textbf{87.58} \\
& & no-edit WER (\%) $\downarrow$ & 3.52~|~17.70 & \textbf{3.51}~|~\textbf{16.56} \\
\cmidrule(lr){2-5}
& \multirow{4}{*}{open}
& WER (\%) $\downarrow$ & 3.89~|~7.59 & \textbf{2.57}~|~\textbf{6.98} \\
& & SIM $\uparrow$ & \textbf{0.83}~|~0.79 & \textbf{0.83}~|~\textbf{0.84} \\
& & ACC (\%) $\uparrow$ & 79.31~|~62.31 & \textbf{86.90}~|~\textbf{69.85} \\
& & no-edit WER (\%) $\downarrow$ & 4.10~|~18.84 & \textbf{2.77}~|~\textbf{17.83} \\
\midrule
\multirow{8}{*}{Substitution}
& \multirow{4}{*}{basic}
& WER (\%) $\downarrow$ & 4.52~|~8.99 & \textbf{2.66}~|~\textbf{4.46} \\
& & SIM $\uparrow$ & 0.82~|~0.78 & \textbf{0.84}~|~\textbf{0.81} \\
& & ACC (\%) $\uparrow$ & 78.62~|~59.78 & \textbf{87.42}~|~\textbf{75.98} \\
& & no-edit WER (\%) $\downarrow$ & 4.63~|~19.28 & \textbf{2.91}~|~\textbf{16.34} \\
\cmidrule(lr){2-5}
& \multirow{4}{*}{open}
& WER (\%) $\downarrow$ & 4.56~|~7.64 & \textbf{2.45}~|~\textbf{4.41} \\
& & SIM $\uparrow$ & 0.83~|~0.77 & \textbf{0.84}~|~\textbf{0.81} \\
& & ACC (\%) $\uparrow$ & 76.62~|~65.62 & \textbf{90.15}~|~\textbf{76.95} \\
& & no-edit WER (\%) $\downarrow$ & 4.75~|~18.39 & \textbf{2.71}~|~\textbf{16.16} \\
\bottomrule
\end{tabular}
\end{table}

\begin{table}[ht]
\centering
\small
\setlength{\tabcolsep}{6pt}
\renewcommand{\arraystretch}{1.10}
\caption{Acoustic-editing results on Ming-Freeform-Audio-Edit. Lower WER, RDE, and RAE, and higher SIM indicate better performance. Best results are shown in bold. Following the benchmark protocol, Pitch Alteration reports only WER and SIM.}
\label{tab:acoustic_speech_editing}
\begin{tabular}{@{}llcc@{}}
\toprule
\begin{tabular}[c]{@{}l@{}}\textbf{Task}\end{tabular}
& \begin{tabular}[c]{@{}l@{}}\textbf{Metric}\end{tabular}
& \begin{tabular}[c]{@{}c@{}}\textbf{Ming-UniAudio-Edit}~\cite{yan2025ming}\\\textbf{(ZH | EN)}\end{tabular}
& \begin{tabular}[c]{@{}c@{}}\textbf{FireRedAudio}\\\textbf{(ZH | EN)}\end{tabular} \\
\midrule
\multirow{3}{*}{Speed Alteration}
& WER (\%) $\downarrow$ & 5.88~|~17.53 & \textbf{2.00}~|~\textbf{4.43} \\
& SIM $\uparrow$ & 0.66~|~0.57 & \textbf{0.79}~|~\textbf{0.71} \\
& RDE (\%) $\downarrow$ & 6.36~|~5.92 & \textbf{2.60}~|~\textbf{4.02} \\
\midrule
\multirow{2}{*}{Pitch Alteration}
& WER (\%) $\downarrow$ & 7.45~|~13.37 & \textbf{2.00}~|~\textbf{3.04} \\
& SIM $\uparrow$ & 0.36~|~0.24 & \textbf{0.52}~|~\textbf{0.44} \\
\midrule
\multirow{3}{*}{Volume Alteration}
& WER (\%) $\downarrow$ & 1.71~|~1.35 & \textbf{1.60}~|~\textbf{1.30} \\
& SIM $\uparrow$ & 0.86~|~0.80 & \textbf{0.94}~|~\textbf{0.93} \\
& RAE (\%) $\downarrow$ & 14.90~|~11.70 & \textbf{2.39}~|~\textbf{3.74} \\
\bottomrule
\end{tabular}
\end{table}

We evaluate free-form speech editing on Ming-Freeform-Audio-Edit~\cite{yan2025ming}, without providing edit locations or alignments. Semantic editing covers deletion, insertion, and substitution under \texttt{basic} and \texttt{open} settings, with \texttt{open} corresponding to the original \texttt{full} setting. Acoustic editing covers speed, pitch, and volume alteration, with 50 Chinese and 50 English examples per task.
Following the benchmark, Chinese CER and English WER are jointly labeled WER. Semantic editing reports overall WER, edit-region accuracy (ACC), no-edit WER for unchanged content, and speaker similarity (SIM). Acoustic editing additionally uses relative duration error (RDE) for speed and relative amplitude error (RAE) for volume; lower error rates and higher ACC/SIM are better. Tables~\ref{tab:semantic_speech_editing} and~\ref{tab:acoustic_speech_editing} report the results.

Tables~\ref{tab:semantic_speech_editing} and~\ref{tab:acoustic_speech_editing} show that FireRedAudio matches or improves upon Ming-UniAudio-Edit on all but one reported value; the sole exception is Chinese SIM under open deletion, with a difference of only 0.01. Across deletion, insertion, and substitution, FireRedAudio consistently reduces both overall and no-edit WER while matching or improving edit-region ACC, with particularly large gains on open deletion. It also improves every reported metric for speed, pitch, and volume alteration. The joint improvements in edit-region accuracy and no-edit WER indicate that FireRedAudio can apply the requested semantic changes while preserving unedited content; the simultaneous gains in RDE or RAE, WER, and SIM show that it can control acoustic attributes without compromising linguistic content or speaker characteristics. This behavior may benefit from using the shared LLM, together with explicit CoT supervision, to interpret the editing instruction and determine what should be modified, while the continuous DiT--RedAE pathway synthesizes the edited waveform.

\subsection{Structured Long-Form Audio Organization with Second-Level Timestamp Accuracy}
\label{sec:long_form_audio_understanding}

\begin{figure*}[ht]
    \centering
    \includegraphics[width=0.9\textwidth]{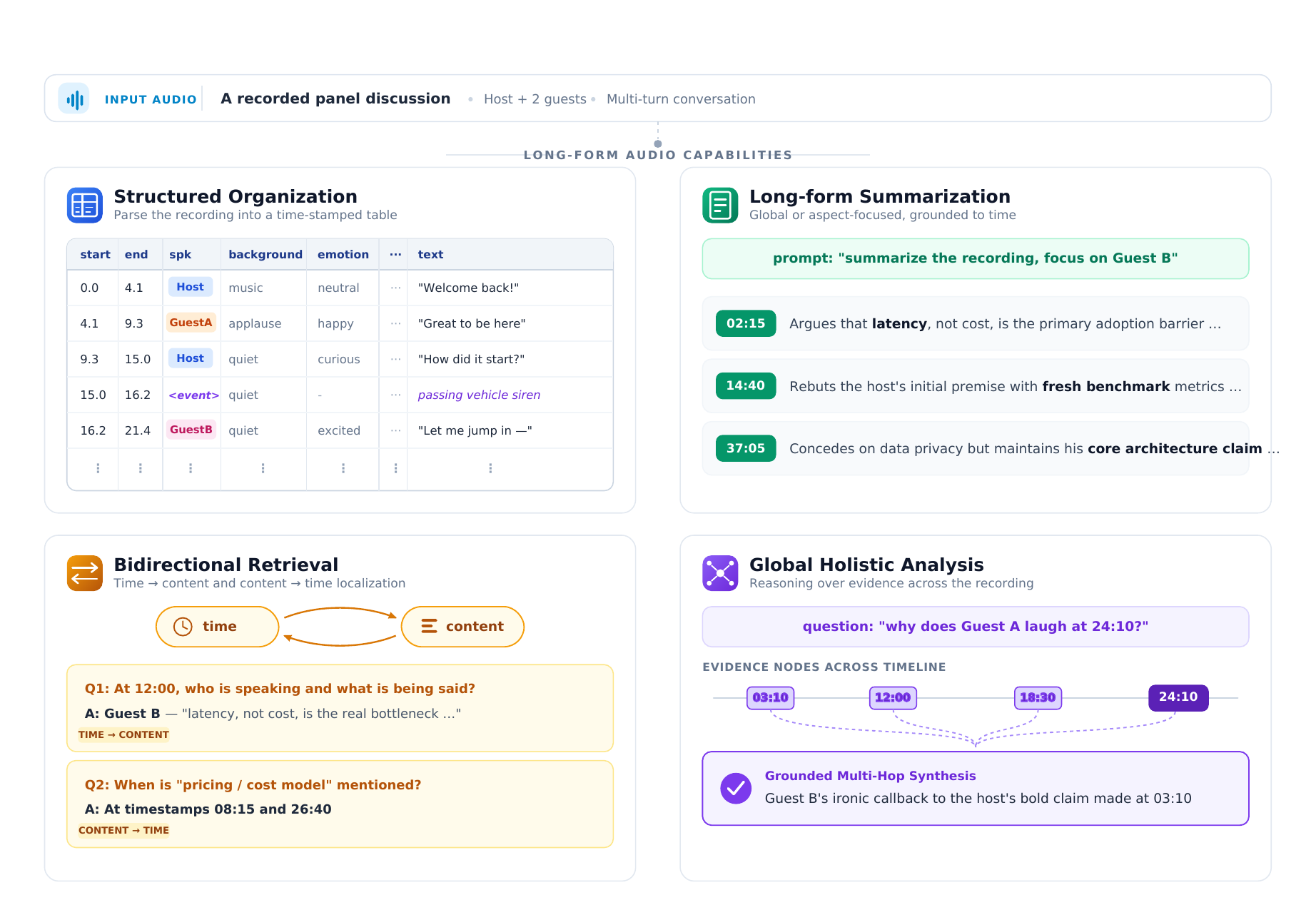}
    \caption{Four representative long-form audio-understanding capabilities: structured organization, long-form summarization, bidirectional retrieval between time and content, and global analysis over evidence distributed across a recording. Our quantitative temporal-grounding evaluation focuses on structured organization.}
    \label{fig:long_audio_capability}
\end{figure*}

FireRedAudio accepts recordings of up to one hour and supports the four representative long-form audio-understanding capabilities illustrated in Figure~\ref{fig:long_audio_capability}. We quantitatively evaluate temporal grounding only for structured organization, in which the model converts a recording into structured entries containing time intervals and corresponding content. This output directly exposes whether the predicted temporal boundaries are consistent with the recorded content.

To quantify temporal grounding in structured organization, we evaluate 50 speech-dominant recordings, with 10 recordings in each of the 5-, 10-, 20-, 30-, and 50-minute duration groups. Each predicted entry contains an interval $[s,e]$ and the corresponding speech content. Because valid structured outputs may segment the same recording differently, we score each predicted entry directly rather than matching it to a fixed reference segmentation. An audio judge determines whether the predicted interval contains the speech content specified by the entry; speaker, emotion, and other non-temporal fields are excluded.
Qwen3.5-Omni-Plus~\cite{team2026qwen3} serves as the audio judge. The same judge and prompt are applied to outputs from both systems. Under \texttt{strict@0}, the exact interval passes only if it contains the complete specified speech content, includes no additional speech, and has no more than $0.3$ seconds of leading or trailing silence. Under \texttt{content@}\(\delta\), the interval is expanded to $[s-\delta,e+\delta]$ and passes if the specified speech content is fully present in the expanded audio segment. We report $\delta=0.5$ and $1.0$ seconds. Higher pass rates indicate better temporal grounding.
The \texttt{Overall} score is computed over all extracted segments from the 50 recordings, rather than by averaging the five duration-group scores.

\begin{table}[ht]
\centering
\footnotesize
\setlength{\tabcolsep}{3pt}
\renewcommand{\arraystretch}{1.10}
\caption{Reference-boundary-free temporal-grounding pass rates (\%) for structured long-form audio organization, grouped by recording duration. Each group contains 10 speech-dominant recordings. Qwen3.5-Omni-Plus is used as the audio judge; higher is better. Best results are shown in bold.}
\label{tab:long_audio_timestamp}
\resizebox{\textwidth}{!}{%
\begin{tabular}{@{}lcccccc@{}}
\toprule
\multirow[c]{2}{*}{\textbf{Duration}}
& \multicolumn{2}{c}{\textbf{strict@0}}
& \multicolumn{2}{c}{\textbf{content@0.5}}
& \multicolumn{2}{c}{\textbf{content@1.0}} \\
\cmidrule(lr){2-3}\cmidrule(lr){4-5}\cmidrule(lr){6-7}
& \textbf{FireRedAudio} & \textbf{Qwen3.5-Omni-Plus}
& \textbf{FireRedAudio} & \textbf{Qwen3.5-Omni-Plus}
& \textbf{FireRedAudio} & \textbf{Qwen3.5-Omni-Plus} \\
\midrule
5 min  & \textbf{76.9} & 45.9 & \textbf{100.0} & 98.1 & \textbf{100.0} & 99.3 \\
10 min & \textbf{68.9} & 53.9 & \textbf{99.0} & 96.0 & \textbf{99.4} & 99.0 \\
20 min & \textbf{73.8} & 61.0 & \textbf{96.8} & 95.6 & 98.0 & \textbf{98.4} \\
30 min & \textbf{77.7} & 71.1 & \textbf{98.4} & 98.2 & \textbf{98.7} & 98.4 \\
50 min & \textbf{72.5} & 47.5 & \textbf{93.9} & 85.5 & \textbf{94.4} & 92.9 \\
\midrule
Overall & \textbf{73.6} & 56.6 & \textbf{96.1} & 92.6 & \textbf{96.7} & 96.5 \\
\bottomrule
\end{tabular}
}
\end{table}

Table~\ref{tab:long_audio_timestamp} shows that FireRedAudio maintains accurate temporal grounding for structured organization from 5- to 50-minute recordings and outperforms Qwen3.5-Omni-Plus on all three overall metrics. Under \texttt{strict@0}, it leads by 17.0 percentage points, indicating more precise and compact interval placement. With a $0.5$-second boundary tolerance, its overall pass rate exceeds 96\% and remains above 93\% in every duration group, including 50-minute recordings. The high \texttt{content@1} pass rate further shows that most structured entries contain their corresponding speech content after the predicted interval is expanded by one second at both ends. Using \texttt{content@1} as the criterion for second-level timestamp accuracy, FireRedAudio maintains high temporal-grounding performance across recordings from 5 to 50 minutes. This robustness may be supported by our long-form data construction, which merges accurately annotated chunks into longer recordings, converts chunk-relative timestamps into recording-level timestamps, and reconciles chunk-level speaker labels into globally consistent speaker identities. Timestamp conversion directly supervises time--content alignment, while speaker-label reconciliation promotes cross-chunk speaker consistency.

\section{Conclusion}
\label{sec:conclusion}

We presented FireRedAudio, a general-purpose audio language model that unifies ASR, audio understanding, zero-shot TTS, Instruct TTS, and semantic and acoustic speech editing within a shared 9B-parameter LLM. Its central design decouples continuous input representations according to the function of each audio segment: a dedicated Audio Encoder serves understanding, while the reconstructible RedAE pathway serves speech generation. Combined with a flow-matching DiT and progressive multitask training, this design enables audio understanding for recordings of up to one hour and achieves competitive or leading performance across the evaluated tasks. For structured organization of long-form audio, FireRedAudio demonstrates second-level timestamp accuracy. Overall, the results support input-side representation decoupling as a viable way to accommodate the different requirements of audio understanding and speech generation while retaining a shared language and reasoning backbone.

\section*{Author Contributions}

Authors are listed alphabetically by given name within each group.

\textbf{Core contributors:} Feiyu Shen, Fenglong Xie, Junjie Li, Kun Xie, Lei Xie, Xu Tang, Xuelong Geng, Yan Jia, Yao Hu, Yichen Han, Yichen Wu, and Ziqi Dai.

\textbf{Contributors:} Junjie Chen, Kai Huang, Manzhen Wei, and Yixuan Li.

\bibliographystyle{unsrt}
\bibliography{refs}

@article{yan2025ming,
  title={Ming-UniAudio: Speech LLM for Joint Understanding, Generation and Editing with Unified Representation},
  author={Yan, Canxiang and Jin, Chunxiang and Huang, Dawei and Yu, Haibing and Peng, Han and Zhan, Hui and Gao, Jie and Peng, Jing and Chen, Jingdong and Zhou, Jun and others},
  journal={arXiv preprint arXiv:2511.05516},
  year={2025}
}

@article{zhou2026f3,
  title={F3-Tokenizer: Taming Audio Autoencoder Latents for Understanding and Generation},
  author={Zhou, Dinghao and Song, Xingchen and Wu, Di and Cheng, Pengyu and Shen, Shengfan and Lv, Sixiang},
  journal={arXiv preprint arXiv:2606.06357},
  year={2026}
}

@article{team2026longcat,
  title={Longcat-next: Lexicalizing modalities as discrete tokens},
  author={Team, Meituan LongCat and Xiao, Bin and Wang, Chao and Li, Chengjiang and Zhang, Chi and Peng, Chong and Yu, Hang and Yang, Hao and Yan, Haonan and Sun, Haoze and others},
  journal={arXiv preprint arXiv:2603.27538},
  year={2026}
}

@article{tian2025ualm,
  title={Ualm: Unified audio language model for understanding, generation and reasoning},
  author={Tian, Jinchuan and Lee, Sang-gil and Kong, Zhifeng and Ghosh, Sreyan and Goel, Arushi and Yang, Chao-Han Huck and Dai, Wenliang and Liu, Zihan and Ye, Hanrong and Watanabe, Shinji and others},
  journal={arXiv preprint arXiv:2510.12000},
  year={2025}
}

@article{kong2026unified,
  title={Unified Audio Intelligence Without Regressing on Text Intelligence},
  author={Kong, Zhifeng and Lee, Sang-gil and Kim, Jaehyeon and Wang, Boxin and Liu, Zihan and Kim, Sungwon and Chen, Yang and Goel, Arushi and Roy, Rajarshi and Dai, Wenliang and others},
  journal={arXiv preprint arXiv:2607.05196},
  year={2026}
}

@article{ding2025kimi,
  title={Kimi-audio technical report},
  author={Ding, Ding and Ju, Zeqian and Leng, Yichong and Liu, Songxiang and Liu, Tong and Shang, Zeyu and Shen, Kai and Song, Wei and Tan, Xu and Tang, Heyi and others},
  journal={arXiv preprint arXiv:2504.18425},
  year={2025}
}

@article{team2026qwen3,
  title={Qwen3.5-omni technical report},
  author={Team, Qwen},
  journal={arXiv preprint arXiv:2604.15804},
  year={2026}
}

@article{xu2025qwen3,
  title={Qwen3-omni technical report},
  author={Xu, Jin and Guo, Zhifang and Hu, Hangrui and Chu, Yunfei and Wang, Xiong and He, Jinzheng and Wang, Yuxuan and Shi, Xian and He, Ting and Zhu, Xinfa and others},
  journal={arXiv preprint arXiv:2509.17765},
  year={2025}
}

@article{yang2026uniaudio,
  title={UniAudio 2.0: A Unified Audio Language Model with Text-Aligned Factorized Audio Tokenization},
  author={Yang, Dongchao and Wang, Yuanyuan and Chong, Dading and Liu, Songxiang and Wu, Xixin and Meng, Helen},
  journal={arXiv preprint arXiv:2602.04683},
  year={2026}
}

@inproceedings{wang2026dualspeechlm,
  title={DualSpeechLM: Towards Unified Speech Understanding and Generation via Dual Speech Token Modeling with Large Language Models},
  author={Wang, Yuanyuan and Yang, Dongchao and Shao, Yiwen and Chen, Hangting and Zhao, Jiankun and Wu, Zhiyong and Meng, Helen and Wu, Xixin},
  booktitle={Proceedings of the AAAI Conference on Artificial Intelligence},
  volume={40},
  pages={33728--33736},
  year={2026}
}

@inproceedings{radford2023robust,
  title={Robust speech recognition via large-scale weak supervision},
  author={Radford, Alec and Kim, Jong Wook and Xu, Tao and Brockman, Greg and McLeavey, Christine and Sutskever, Ilya},
  booktitle={International conference on machine learning},
  pages={28492--28518},
  year={2023},
  organization={PMLR}
}

@article{sun2024multimodal,
  title={Multimodal latent language modeling with next-token diffusion},
  author={Sun, Yutao and Bao, Hangbo and Wang, Wenhui and Peng, Zhiliang and Dong, Li and Huang, Shaohan and Wang, Jianyong and Wei, Furu},
  journal={arXiv preprint arXiv:2412.08635},
  year={2024}
}

@article{jia2025ditar,
  title={Ditar: Diffusion transformer autoregressive modeling for speech generation},
  author={Jia, Dongya and Chen, Zhuo and Chen, Jiawei and Du, Chenpeng and Wu, Jian and Cong, Jian and Zhuang, Xiaobin and Li, Chumin and Wei, Zhen and Wang, Yuping and others},
  journal={arXiv preprint arXiv:2502.03930},
  year={2025}
}

@article{peng2025vibevoice,
  title={Vibevoice technical report},
  author={Peng, Zhiliang and Yu, Jianwei and Wang, Wenhui and Chang, Yaoyao and Sun, Yutao and Dong, Li and Zhu, Yi and Xu, Weijiang and Bao, Hangbo and Wang, Zehua and others},
  journal={arXiv preprint arXiv:2508.19205},
  year={2025}
}

@article{ai2025ming,
  title={Ming-flash-omni: A sparse, unified architecture for multimodal perception and generation},
  author={AI, Inclusion and Ma, Bowen and Zou, Cheng and Du, ChengKun and Yan, Canxiang and Jin, Chunxiang and Shen, Chunjie and Lian, Chenyu and Fan, Chengxiang and Zheng, Dandan and others},
  journal={arXiv preprint arXiv:2510.24821},
  year={2025}
}

@article{tian2026audio,
  title={Audio-Omni: Extending Multi-modal Understanding to Versatile Audio Generation and Editing},
  author={Tian, Zeyue and Yang, Binxin and Liu, Zhaoyang and Zhang, Jiexuan and Yuan, Ruibin and Yin, Hubery and Chen, Qifeng and Li, Chen and Lyu, Jing and Xue, Wei and others},
  journal={arXiv preprint arXiv:2604.10708},
  year={2026}
}

@article{wang2026uat,
  title={UAT: Unified Audio-Text Diffusion for Audio Generation, Editing, and Captioning},
  author={Wang, Hui and Yang, Yifan and Tian, Zeyue and Jia, Yuhang and Zhao, Jinghua and Zhou, Long and Han, Bing and Liu, Cheng and Zhou, Jiaming and Tu, Geng and others},
  journal={arXiv preprint arXiv:2606.04939},
  year={2026}
}

@inproceedings{peebles2023scalable,
  title={Scalable diffusion models with transformers},
  author={Peebles, William and Xie, Saining},
  booktitle={Proceedings of the IEEE/CVF international conference on computer vision},
  pages={4195--4205},
  year={2023}
}

@misc{gemini31_pro,
  author = {{Google DeepMind}},
  title  = {{Gemini 3.1 Pro Model Card}},
  year   = {2026},
  month  = feb,
  url    = {https://storage.googleapis.com/deepmind-media/Model-Cards/Gemini-3-1-Pro-Model-Card.pdf},
  note   = {Published February 19, 2026; accessed July 20, 2026}
}

@article{kingma2013auto,
  title={Auto-encoding variational bayes},
  author={Kingma, Diederik P and Welling, Max},
  journal={arXiv preprint arXiv:1312.6114},
  year={2013}
}

@article{lipman2022flow,
  title={Flow matching for generative modeling},
  author={Lipman, Yaron and Chen, Ricky TQ and Ben-Hamu, Heli and Nickel, Maximilian and Le, Matt},
  journal={arXiv preprint arXiv:2210.02747},
  year={2022}
}

@inproceedings{sakshi2025mmau,
  title={Mmau: A massive multi-task audio understanding and reasoning benchmark},
  author={Sakshi, Sakshi and Tyagi, Utkarsh and Kumar, Sonal and Seth, Ashish and Selvakumar, Ramaneswaran and Nieto, Oriol and Duraiswami, Ramani and Ghosh, Sreyan and Manocha, Dinesh},
  booktitle={International Conference on Learning Representations},
  volume={2025},
  pages={84929--84964},
  year={2025}
}

@article{wang2025mmsu,
  title={Mmsu: A massive multi-task spoken language understanding and reasoning benchmark},
  author={Wang, Dingdong and Li, Junan and Wu, Jincenzi and Yang, Dongchao and Chen, Xueyuan and Zhang, Tianhua and Meng, Helen},
  journal={arXiv preprint arXiv:2506.04779},
  year={2025}
}

@article{anastassiou2024seed,
  title={Seed-tts: A family of high-quality versatile speech generation models},
  author={Anastassiou, Philip and Chen, Jiawei and Chen, Jitong and Chen, Yuanzhe and Chen, Zhuo and Chen, Ziyi and Cong, Jian and Deng, Lelai and Ding, Chuang and Gao, Lu and others},
  journal={arXiv preprint arXiv:2406.02430},
  year={2024}
}

@article{huang2025instructttseval,
  title={Instructttseval: Benchmarking complex natural-language instruction following in text-to-speech systems},
  author={Huang, Kexin and Tu, Qian and Fan, Liwei and Yang, Chenchen and Zhang, Dong and Li, Shimin and Fei, Zhaoye and Cheng, Qinyuan and Qiu, Xipeng},
  journal={arXiv preprint arXiv:2506.16381},
  year={2025}
}

@article{yang2025qwen3,
  title={Qwen3 technical report},
  author={Yang, An and Li, Anfeng and Yang, Baosong and Zhang, Beichen and Hui, Binyuan and Zheng, Bo and Yu, Bowen and Gao, Chang and Huang, Chengen and Lv, Chenxu and others},
  journal={arXiv preprint arXiv:2505.09388},
  year={2025}
}

@inproceedings{ye2025codec,
  title={Codec does matter: Exploring the semantic shortcoming of codec for audio language model},
  author={Ye, Zhen and Sun, Peiwen and Lei, Jiahe and Lin, Hongzhan and Tan, Xu and Dai, Zheqi and Kong, Qiuqiang and Chen, Jianyi and Pan, Jiahao and Liu, Qifeng and others},
  booktitle={Proceedings of the AAAI Conference on Artificial Intelligence},
  volume={39},

  pages={25697--25705},
  year={2025}
}

@inproceedings{siuzdak2024vocos,
  title={Vocos: Closing the gap between time-domain and fourier-based neural vocoders for high-quality audio synthesis},
  author={Siuzdak, Hubert},
  booktitle={International Conference on Learning Representations},
  volume={2024},
  pages={25719--25733},
  year={2024}
}

@article{ho2022classifier,
  title={Classifier-free diffusion guidance},
  author={Ho, Jonathan and Salimans, Tim},
  journal={arXiv preprint arXiv:2207.12598},
  year={2022}
}

@inproceedings{bu2017aishell,
  title={Aishell-1: An open-source mandarin speech corpus and a speech recognition baseline},
  author={Bu, Hui and Du, Jiayu and Na, Xingyu and Wu, Bengu and Zheng, Hao},
  booktitle={2017 20th conference of the oriental chapter of the international coordinating committee on speech databases and speech I/O systems and assessment (O-COCOSDA)},
  pages={1--5},
  year={2017},
  organization={IEEE}
}

@article{du2018aishell,
  title={Aishell-2: Transforming mandarin asr research into industrial scale},
  author={Du, Jiayu and Na, Xingyu and Liu, Xuechen and Bu, Hui},
  journal={arXiv preprint arXiv:1808.10583},
  year={2018}
}

@inproceedings{zhang2022wenetspeech,
  title={Wenetspeech: A 10000+ hours multi-domain mandarin corpus for speech recognition},
  author={Zhang, Binbin and Lv, Hang and Guo, Pengcheng and Shao, Qijie and Yang, Chao and Xie, Lei and Xu, Xin and Bu, Hui and Chen, Xiaoyu and Zeng, Chenchen and others},
  booktitle={ICASSP 2022-2022 IEEE International Conference on Acoustics, Speech and Signal Processing (ICASSP)},
  pages={6182--6186},
  year={2022},
  organization={IEEE}
}

@inproceedings{panayotov2015librispeech,
  title={Librispeech: an asr corpus based on public domain audio books},
  author={Panayotov, Vassil and Chen, Guoguo and Povey, Daniel and Khudanpur, Sanjeev},
  booktitle={2015 IEEE international conference on acoustics, speech and signal processing (ICASSP)},
  pages={5206--5210},
  year={2015},
  organization={IEEE}
}

@article{wang2022opencpop,
  title={Opencpop: A high-quality open source chinese popular song corpus for singing voice synthesis},
  author={Wang, Yu and Wang, Xinsheng and Zhu, Pengcheng and Wu, Jie and Li, Hanzhao and Xue, Heyang and Zhang, Yongmao and Xie, Lei and Bi, Mengxiao},
  journal={arXiv preprint arXiv:2201.07429},
  year={2022}
}

@article{fleurs2022arxiv,
  title = {FLEURS: Few-shot Learning Evaluation of Universal Representations of Speech},
  author = {Conneau, Alexis and Ma, Min and Khanuja, Simran and Zhang, Yu and Axelrod, Vera and Dalmia, Siddharth and Riesa, Jason and Rivera, Clara and Bapna, Ankur},
  journal={arXiv preprint arXiv:2205.12446},
  url = {https://arxiv.org/abs/2205.12446},
  year = {2022},
}

@inproceedings{tang2021kespeech,
  title={Kespeech: An open source speech dataset of mandarin and its eight subdialects},
  author={Tang, Zhiyuan and Wang, Dong and Xu, Yanguang and Sun, Jianwei and Lei, Xiaoning and Zhao, Shuaijiang and Wen, Cheng and Tan, Xingjun and Xie, Chuandong and Zhou, Shuran and others},
  booktitle={Thirty-fifth conference on neural information processing systems datasets and benchmarks track (Round 2)},
  year={2021}
}

@article{tian2025step,
  title={Step-Audio-R1 Technical Report},
  author={Tian, Fei and Zhang, Xiangyu Tony and Zhang, Yuxin and Zhang, Haoyang and Li, Yuxin and Liu, Daijiao and Deng, Yayue and Wu, Donghang and Chen, Jun and Zhao, Liang and others},
  journal={arXiv preprint arXiv:2511.15848},
  year={2025}
}

@article{zhang2025mimo,
  title={MiMo-Audio: Audio Language Models are Few-Shot Learners},
  author={Zhang, Dong and Wang, Gang and Xue, Jinlong and Fang, Kai and Zhao, Liang and Ma, Rui and Ren, Shuhuai and Liu, Shuo and Guo, Tao and Zhuang, Weiji and others},
  journal={arXiv preprint arXiv:2512.23808},
  year={2025}
}

@inproceedings{ardila2020common,
  title={Common voice: A massively-multilingual speech corpus},
  author={Ardila, Rosana and Branson, Megan and Davis, Kelly and Kohler, Michael and Meyer, Josh and Henretty, Michael and Morais, Reuben and Saunders, Lindsay and Tyers, Francis and Weber, Gregor},
  booktitle={Proceedings of the twelfth language resources and evaluation conference},
  pages={4218--4222},
  year={2020}
}

@inproceedings{guo2021didispeech,
  title={Didispeech: A large scale mandarin speech corpus},
  author={Guo, Tingwei and Wen, Cheng and Jiang, Dongwei and Luo, Ne and Zhang, Ruixiong and Zhao, Shuaijiang and Li, Wubo and Gong, Cheng and Zou, Wei and Han, Kun and others},
  booktitle={ICASSP 2021-2021 IEEE International Conference on Acoustics, Speech and Signal Processing (ICASSP)},
  pages={6968--6972},
  year={2021},
  organization={IEEE}
}

@inproceedings{gao2022paraformer,
  title={Paraformer: Fast and Accurate Parallel Transformer for Non-autoregressive End-to-End Speech Recognition},
  author={Gao, Zhifu and Zhang, Shiliang and McLoughlin, Ian and Yan, Zhijie},
  booktitle={INTERSPEECH},
  year={2022}
}

@article{chen2022wavlm,
  title={Wavlm: Large-scale self-supervised pre-training for full stack speech processing},
  author={Chen, Sanyuan and Wang, Chengyi and Chen, Zhengyang and Wu, Yu and Liu, Shujie and Chen, Zhuo and Li, Jinyu and Kanda, Naoyuki and Yoshioka, Takuya and Xiao, Xiong and others},
  journal={IEEE Journal of Selected Topics in Signal Processing},
  volume={16},
  number={6},
  pages={1505--1518},
  year={2022},
  publisher={IEEE}
}

@article{guo2024fireredtts,
  title={Fireredtts: A foundation text-to-speech framework for industry-level generative speech applications},
  author={Guo, Hao-Han and Hu, Yao and Liu, Kun and Shen, Fei-Yu and Tang, Xu and Wu, Yi-Chen and Xie, Feng-Long and Xie, Kun and Xu, Kai-Tuo},
  journal={arXiv preprint arXiv:2409.03283},
  year={2024}
}

@article{xie2025fireredtts,
  title={Fireredtts-2: Towards long conversational speech generation for podcast and chatbot},
  author={Xie, Kun and Shen, Feiyu and Li, Junjie and Xie, Fenglong and Tang, Xu and Hu, Yao},
  journal={arXiv preprint arXiv:2509.02020},
  year={2025}
}

@inproceedings{chen2025f5,
  title={F5-tts: A fairytaler that fakes fluent and faithful speech with flow matching},
  author={Chen, Yushen and Niu, Zhikang and Ma, Ziyang and Deng, Keqi and Wang, Chunhui, JianZhao and Yu, Kai and Chen, Xie},
  booktitle={Proceedings of the 63rd Annual Meeting of the Association for Computational Linguistics (Volume 1: Long Papers)},
  pages={6255--6271},
  year={2025}
}

@article{du2024cosyvoice,
  title={Cosyvoice 2: Scalable streaming speech synthesis with large language models},
  author={Du, Zhihao and Wang, Yuxuan and Chen, Qian and Shi, Xian and Lv, Xiang and Zhao, Tianyu and Gao, Zhifu and Yang, Yexin and Gao, Changfeng and Wang, Hui and others},
  journal={arXiv preprint arXiv:2412.10117},
  year={2024}
}

@article{du2025cosyvoice,
  title={Cosyvoice 3: Towards in-the-wild speech generation via scaling-up and post-training},
  author={Du, Zhihao and Gao, Changfeng and Wang, Yuxuan and Yu, Fan and Zhao, Tianyu and Wang, Hao and Lv, Xiang and Wang, Hui and Ni, Chongjia and Shi, Xian and others},
  journal={arXiv preprint arXiv:2505.17589},
  year={2025}
}

@article{Qwen2.5-Omni, title={Qwen2.5-Omni Technical Report}, author={Jin Xu and Zhifang Guo and Jinzheng He and Hangrui Hu and Ting He and Shuai Bai and Keqin Chen and Jialin Wang and Yang Fan and Kai Dang and Bin Zhang and Xiong Wang and Yunfei Chu and Junyang Lin}, journal={arXiv preprint arXiv:2503.20215}, year={2025} }

@misc{gemini25_pro,
  author = {{Google DeepMind}},
  title  = {{Gemini 2.5 Pro Model Card}},
  year   = {2025},
  month  = jun,
  url    = {https://storage.googleapis.com/deepmind-media/Model-Cards/Gemini-2-5-Pro-Model-Card.pdf},
  note   = {Updated June 27, 2025; accessed July 20, 2026}
}

@article{huang2026moss,
  title={MOSS-VoiceGenerator: Create Realistic Voices with Natural Language Descriptions},
  author={Huang, Kexin and Fan, Liwei and Jiang, Botian and Jiang, Yaozhou and Tu, Qian and Zhu, Jie and Zhang, Yuqian and Zhao, Yiwei and Yang, Chenchen and Fei, Zhaoye and others},
  journal={arXiv preprint arXiv:2603.28086},
  year={2026}
}

@misc{mingomnitts,  
  author={Inclusion AI},
  title={Ming-omni-tts: Simple and Efficient Unified Generation of Speech, Music, and Sound with Precise Control},
  year={2026},
  url={https://github.com/inclusionAI/Ming-omni-tts},
  note={[Accessed 2026-04-01]}
}

@article{hu2026qwen3,
  title={Qwen3-TTS Technical Report},
  author={Hu, Hangrui and Zhu, Xinfa and He, Ting and Guo, Dake and Zhang, Bin and Wang, Xiong and Guo, Zhifang and Jiang, Ziyue and Hao, Hongkun and Guo, Zishan and others},
  journal={arXiv preprint arXiv:2601.15621},
  year={2026}
}

@article{wu2025step,
  title={Step-audio 2 technical report},
  author={Wu, Boyong and Yan, Chao and Hu, Chen and Yi, Cheng and Feng, Chengli and Tian, Fei and Shen, Feiyu and Yu, Gang and Zhang, Haoyang and Li, Jingbei and others},
  journal={arXiv preprint arXiv:2507.16632},
  year={2025}
}

@incollection{joyce2025kullback,
  title={Kullback-leibler divergence},
  author={Joyce, James M},
  booktitle={International encyclopedia of statistical science},
  pages={1307--1309},
  year={2025},
  publisher={Springer}
}

@article{conneau2020unsupervised,
  title={Unsupervised cross-lingual representation learning for speech recognition},
  author={Conneau, Alexis and Baevski, Alexei and Collobert, Ronan and Mohamed, Abdelrahman and Auli, Michael},
  journal={arXiv preprint arXiv:2006.13979},
  year={2020}
}

@article{hu2026voicesculptor,
  title={Voicesculptor: Your voice, designed by you},
  author={Hu, Jingbin and Chen, Huakang and Ma, Linhan and Guo, Dake and Zhan, Qirui and Li, Wenhao and Zhang, Haoyu and Xia, Kangxiang and Zhang, Ziyu and Tian, Wenjie and others},
  journal={arXiv preprint arXiv:2601.10629},
  year={2026}
}

\clearpage
\appendix
\renewcommand{\thesection}{Appendix~\Alph{section}}

\section{Detailed Multilingual ASR Results}
\label{app:multilingual_asr_results}

Table~\ref{tab:fleurs102} reports the per-language results underlying the FLEURS-102 macro-average in Table~\ref{tab:asr_results}, using a scoring unit appropriate to each writing system.
Depending on the writing system, we report WER for most languages, CER for Chinese, Japanese, and Cantonese, space-stripped CER (sCER) for Korean, and grapheme-cluster error rate (GCER) for Lao, Burmese, Thai, and Khmer. GCER treats each extended grapheme cluster as one scoring unit.

\begin{table}[!ht]
\centering
\caption{Per-language ASR error rates (\%) on FLEURS-102 for FireRedAudio, Qwen3.5-Omni-Plus, and Gemini 3.1 Pro. Lower is better; the best system per language is in bold. Unmarked values use WER; $^{\dagger}$ denotes CER, $^{\ddagger}$ denotes sCER, and $^{\S}$ denotes GCER.}
\label{tab:fleurs102}
\setlength{\tabcolsep}{3.0pt}
\renewcommand{\arraystretch}{0.95}
\footnotesize
\resizebox{\textwidth}{!}{%
\begin{tabular}{@{}lrrr@{\hspace{10pt}}lrrr@{}}
\toprule
\textbf{Language}
& \textbf{FireRedAudio}
& \textbf{Qwen3.5-Omni-Plus}
& \textbf{Gemini 3.1 Pro}
& \textbf{Language}
& \textbf{FireRedAudio}
& \textbf{Qwen3.5-Omni-Plus}
& \textbf{Gemini 3.1 Pro} \\
\midrule
Afrikaans           & 15.71          & \textbf{12.19} & 15.24          & Ganda             & \textbf{24.06} & 50.96          & 43.30          \\
Amharic             & 29.18          & 93.83          & \textbf{17.50} & Lingala           & \textbf{10.31} & 28.40          & 25.61          \\
Arabic              & 7.18           & \textbf{4.00}  & 6.53           & Lao$^{\S}$        & \textbf{17.65} & 58.90          & 19.22          \\
Assamese            & \textbf{12.63} & 27.71          & 18.88          & Lithuanian        & 13.93          & 12.61          & \textbf{6.76}  \\
Asturian            & \textbf{12.38} & 17.50          & 32.39          & Luo               & \textbf{28.99} & 59.44          & 39.99          \\
Azerbaijani         & 16.70          & 7.66           & \textbf{6.91}  & Latvian           & 6.91           & 7.83           & \textbf{4.61}  \\
Belarusian          & \textbf{5.63}  & 7.12           & 6.24           & M\={a}ori         & \textbf{17.81} & 26.89          & 33.13          \\
Bulgarian           & 9.69           & 7.15           & \textbf{5.61}  & Macedonian        & 6.53           & \textbf{4.66}  & 5.03           \\
Bengali             & \textbf{4.88}  & 12.74          & 12.54          & Malayalam         & \textbf{9.61}  & 19.42          & 19.45          \\
Bosnian             & 8.25           & \textbf{5.44}  & 7.15           & Mongolian         & 20.79          & 16.87          & \textbf{13.74} \\
Catalan             & \textbf{1.48}  & 3.42           & 3.53           & Marathi           & \textbf{10.36} & 17.01          & 14.86          \\
Cebuano             & 12.32          & \textbf{10.57} & 13.71          & Malay             & 6.12           & \textbf{3.15}  & 4.49           \\
Sorani Kurdish      & 32.96          & 38.09          & \textbf{29.33} & Maltese           & 17.90          & 24.84          & \textbf{15.74} \\
Chinese$^{\dagger}$ & 3.14           & \textbf{2.46}  & 4.28           & Burmese$^{\S}$    & 18.26          & 68.73          & \textbf{9.21}  \\
Czech               & 6.54           & \textbf{2.67}  & 4.08           & Norwegian         & 8.23           & \textbf{3.81}  & 5.63           \\
Welsh               & \textbf{13.41} & 46.68          & 19.06          & Nepali            & \textbf{12.46} & 20.53          & 22.44          \\
Danish              & 9.79           & \textbf{3.87}  & 6.27           & Dutch             & 4.98           & \textbf{2.91}  & 4.05           \\
German              & 3.00           & \textbf{2.25}  & 3.23           & Northern Sotho    & \textbf{17.00} & 80.87          & 31.46          \\
Greek               & 10.81          & \textbf{4.62}  & 5.59           & Nyanja            & \textbf{21.82} & 65.22          & 41.37          \\
English             & \textbf{2.53}  & 3.33           & 2.97           & Occitan           & \textbf{16.86} & 55.67          & 38.03          \\
Spanish             & 2.00           & \textbf{1.77}  & 2.48           & Oromo             & \textbf{67.31} & 69.41          & 67.83          \\
Estonian            & 8.60           & 7.56           & \textbf{5.40}  & Odia              & \textbf{11.40} & 23.61          & 17.50          \\
Persian             & 10.96          & \textbf{8.61}  & 12.19          & Punjabi           & \textbf{12.77} & 14.97          & 13.41          \\
Fula                & \textbf{45.43} & 80.10          & 67.60          & Polish            & 5.21           & \textbf{2.21}  & 3.31           \\
Finnish             & 9.75           & \textbf{2.61}  & 3.50           & Pashto            & 41.19          & 38.79          & \textbf{37.40} \\
Filipino            & 8.38           & \textbf{5.35}  & 8.16           & Portuguese        & 3.02           & \textbf{2.28}  & 3.19           \\
French              & 3.44           & \textbf{2.23}  & 3.53           & Romanian          & 7.48           & \textbf{3.46}  & 3.80           \\
Irish               & \textbf{47.47} & 76.62          & 74.99          & Russian           & 4.25           & \textbf{2.95}  & 3.60           \\
Galician            & \textbf{4.56}  & 4.66           & 4.79           & Sindhi            & 23.12          & 29.56          & \textbf{18.86} \\
Gujarati            & \textbf{8.82}  & 14.80          & 14.67          & Slovak            & 6.19           & 4.09           & \textbf{4.01}  \\
Hausa               & 26.54          & 31.87          & \textbf{24.81} & Slovenian         & 9.70           & \textbf{6.88}  & 7.56           \\
Hebrew              & 21.80          & \textbf{12.53} & 15.94          & Shona             & \textbf{18.68} & 53.76          & 36.51          \\
Hindi               & \textbf{6.80}  & 8.06           & 10.53          & Somali            & 48.40          & 60.74          & \textbf{42.37} \\
Croatian            & 8.10           & \textbf{5.63}  & 5.93           & Serbian           & 17.31          & 29.22          & \textbf{15.45} \\
Hungarian           & 10.42          & \textbf{5.57}  & 7.21           & Swedish           & 6.68           & \textbf{3.43}  & 5.15           \\
Armenian            & 10.87          & 22.31          & \textbf{6.60}  & Swahili           & \textbf{4.47}  & 10.17          & 10.28          \\
Indonesian          & 2.49           & \textbf{2.06}  & 3.17           & Tamil             & \textbf{14.32} & 22.79          & 24.61          \\
Igbo                & \textbf{37.18} & 85.75          & 58.46          & Telugu            & \textbf{15.73} & 21.06          & 19.96          \\
Icelandic           & 21.78          & \textbf{4.49}  & 6.05           & Tajik             & 14.82          & \textbf{10.33} & 41.34          \\
Italian             & 1.66           & \textbf{1.07}  & 1.61           & Thai$^{\S}$       & 6.16           & \textbf{3.33}  & 5.16           \\
Japanese$^{\dagger}$ & 3.21           & \textbf{2.14}  & 3.13           & Turkish           & 6.11           & \textbf{3.14}  & 4.10           \\
Javanese            & 14.68          & \textbf{12.91} & 14.22          & Ukrainian         & 5.72           & \textbf{3.35}  & 3.86           \\
Georgian            & 15.28          & 25.31          & \textbf{7.27}  & Umbundu           & \textbf{46.67} & 103.54         & 86.89          \\
Kamba               & \textbf{45.95} & 89.78          & 66.21          & Urdu              & \textbf{9.05}  & 15.34          & 17.64          \\
Kabuverdianu        & \textbf{13.81} & 56.10          & 45.65          & Uzbek             & 16.66          & 16.14          & \textbf{12.80} \\
Kazakh              & 10.40          & \textbf{6.28}  & 6.56           & Vietnamese        & 3.53           & \textbf{2.39}  & 3.20           \\
Khmer$^{\S}$        & 21.21          & 30.04          & \textbf{9.82}  & Wolof             & \textbf{28.58} & 46.99          & 40.37          \\
Kannada             & \textbf{9.94}  & 17.80          & 17.70          & Xhosa             & \textbf{25.82} & 65.00          & 37.01          \\
Korean$^{\ddagger}$ & 4.87           & \textbf{3.74}  & 4.33           & Yoruba            & \textbf{36.88} & 76.94          & 69.34          \\
Kyrgyz              & 12.50          & 10.44          & \textbf{10.42} & Cantonese$^{\dagger}$ & 3.13           & \textbf{2.76}  & 6.85           \\
Luxembourgish       & \textbf{26.94} & 36.59          & 31.21          & Zulu              & \textbf{18.68} & 48.08          & 22.75          \\
\bottomrule
\end{tabular}%
}
\end{table}

\end{document}